\documentclass[10pt,journal,compsoc]{IEEEtran}
\ifCLASSOPTIONcompsoc
  \usepackage[nocompress]{cite}
\else
  \usepackage{cite}
\fi
\ifCLASSINFOpdf
  \usepackage[pdftex]{graphicx}
\else
\fi
\usepackage{amsmath}
\ifCLASSOPTIONcompsoc
 \usepackage[caption=false,font=footnotesize,labelfont=sf,textfont=sf]{subfig}
\else
 \usepackage[caption=false,font=footnotesize]{subfig}
\fi
\usepackage[pagebackref=true,breaklinks=true,letterpaper=true,colorlinks,bookmarks=false]{hyperref}
\usepackage{color}
\usepackage{ragged2e}  % use justify
\usepackage{amssymb}
\usepackage{bm}
\newcommand{\eg}{\textit{e.g.},}
\newcommand{\ie}{\textit{i.e.}}
\newcommand{\etal}{\textit{et al.}}
\DeclareMathOperator*{\argmin}{argmin}

\begin{document}
\title{GLEAN: Generative Latent Bank for Image Super-Resolution and Beyond}

\author{Kelvin C.K. Chan, Xiangyu Xu, Xintao Wang, Jinwei Gu,~\IEEEmembership{Senior Member,~IEEE}\\Chen Change Loy,~\IEEEmembership{Senior Member,~IEEE} % <-this % stops a space
  \IEEEcompsocitemizethanks{\IEEEcompsocthanksitem K. C. K. Chan, X. Xu, and C. C. Loy are with S-Lab, Nanyang Technological University (NTU), Singapore (E-mail: chan0899@ntu.edu.sg, xiangyu.xu@ntu.edu.sg, ccloy@ntu.edu.sg).
    \IEEEcompsocthanksitem X. Wang is with Applied Research Center, Tencent PCG (Email: xintao.wang@outlook.com).
    \IEEEcompsocthanksitem J. Gu is with Tetras. AI. and Shanghai AI Laboratory (Email: gujinwei@tetras.ai).
    \IEEEcompsocthanksitem C. C. Loy is the corresponding author.}
  %	\thanks{Manuscript received April 19, 2005; revised August 26, 2021.}
}


\markboth{IEEE TRANSACTIONS ON PATTERN ANALYSIS AND MACHINE INTELLIGENCES}%
{Shell \MakeLowercase{\textit{et al.}}: Bare Demo of IEEEtran.cls for Computer Society Journals}



\IEEEtitleabstractindextext{%
  \justify  % crucial to justify the abstract
  \begin{abstract}
    We show that pre-trained Generative Adversarial Networks (GANs) such as StyleGAN and BigGAN can be used as a latent bank to improve the performance of image super-resolution.
    While most existing perceptual-oriented approaches attempt to generate realistic outputs through learning with adversarial loss, our method, \textbf{G}enerative \textbf{L}at\textbf{E}nt b\textbf{AN}k (GLEAN), goes beyond existing practices by directly leveraging rich and diverse priors encapsulated in a pre-trained GAN.
    But unlike prevalent GAN inversion methods that require expensive image-specific optimization at runtime, our approach only needs a single forward pass for restoration.
    GLEAN can be easily incorporated in a simple encoder-bank-decoder architecture with multi-resolution skip connections. Employing priors from different generative models allows GLEAN to be applied to diverse categories (\eg~human faces, cats, buildings, and cars).
    We further present a lightweight version of GLEAN, named LightGLEAN, which retains only the critical components in GLEAN. Notably, LightGLEAN consists of only 21\% of parameters and 35\% of FLOPs while achieving comparable image quality. We extend our method to different tasks including image colorization and blind image restoration, and extensive experiments show that our proposed models perform favorably in comparison to existing methods. Codes and models are available at https://github.com/open-mmlab/mmediting.
  \end{abstract}

  }

\maketitle

\IEEEdisplaynontitleabstractindextext

% \fi
\IEEEpeerreviewmaketitle

% \IEEEraisesectionheading{\section{Introduction}\label{sec:introduction}}


\begin{figure*}[!t]
  \begin{center}
    \includegraphics[width=0.97\textwidth]{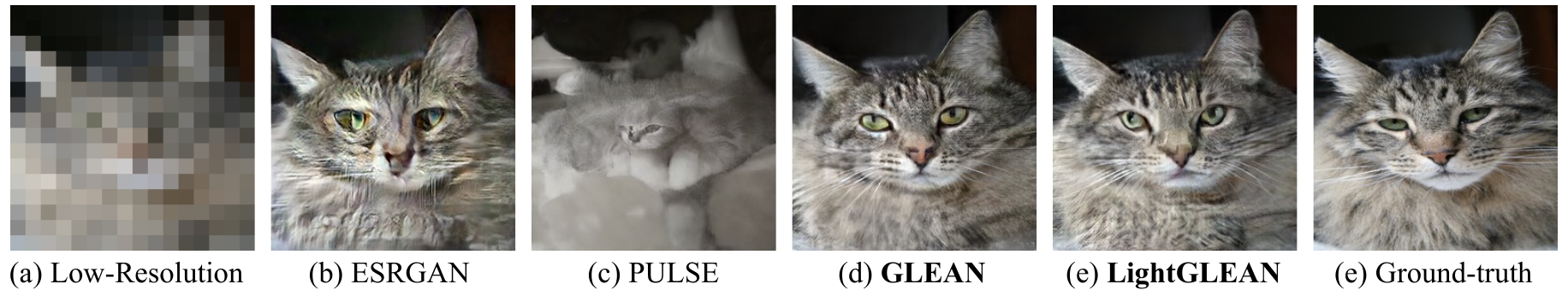}
    \vspace{-0.2cm}
    \caption{\textbf{Example of 16${\times}$ super-resolution (SR)}. (a) The low-resolution input.
      (b) ESRGAN~\cite{wang2018esrgan} trains the SR generator from scratch, often produces artifacts and unnatural textures.
      (c) PULSE~\cite{menon2020pulse}  achieves more realistic textures through GAN inversion but fails to recover ground-truth structures.
      (d) With the proposed generative latent bank, GLEAN is able to generate output that not only is close to the ground-truth, but also possesses realistic textures.
      (e) Our lightweight model, LightGLEAN, achieves comparable performance while having significantly fewer parameters.
      (f) The ground-truth image.
    }
    \label{fig:teaser}
    \vspace{-0.4cm}
  \end{center}
\end{figure*}

% Body Text
% !TEX root = ../submission.tex
% \vspace{-0.3cm}
\IEEEraisesectionheading{\section{Introduction}\label{sec:introduction}}

\IEEEPARstart{I}{n} this study, we explore a new way to employ GAN~\cite{goodfellow2014generative} for image super-resolution.
Since the task of super-resolution is severely underspecified, strong priors are usually required to regularize the restoration process, and the generative prior of GANs has become one of the most widely-used priors thanks to its remarkable abilities to approximate the natural image manifold and synthesize high-quality images.

There are two popular approaches to deploy GANs for super-resolution.
The more common paradigm~\cite{ledig2017photo,wang2018esrgan,wang2018recovering} trains a generator to handle the restoration, where adversarial training is performed by using a discriminator to differentiate real images from the upscaled images produced by the generator.
Another way to exploit GAN is GAN inversion~\cite{bau2020semantic,gu2020image,menon2020pulse,pang2020exploiting}, which needs to `invert' the generation process of a pre-trained GAN by mapping an image back to the latent space. A restored image can then be reconstructed from the optimal vector in the latent space.

While both methods are capable of generating more realistic results than those approaches that solely rely on pixel-wise loss, they have some inherent shortcomings.
The first paradigm typically trains the generator \textit{from scratch} using a combined objective function consisting of an adversarial loss and a fidelity loss.
In this setting, the generator is responsible for both capturing natural image characteristics and maintaining fidelity to the ground truth. This inevitably limits the capability of approximating the natural image manifold. As a result, these methods often produce artifacts, such as unnatural textures and colors.
As shown in Fig.~\ref{fig:teaser}(b), while ESRGAN~\cite{wang2018esrgan} faithfully recovers the structures (\eg~pose, ear shape) of the cat, it struggles to produce realistic textures.

The second paradigm resolves the aforementioned problem by making better use of the latent space of GAN through optimization. However, because the low-dimensional latent codes and the constraints in the image space are insufficient in guiding the restoration process, these methods often generate images with low fidelity.
As shown in Fig.~\ref{fig:teaser}(c), although PULSE~\cite{menon2020pulse} successfully produces a cat-like object, the GAN-inversion-based method fails to recover the structures of the ground-truth faithfully.
In addition, since the optimization is usually conducted in an iterative manner for each image at runtime, these approaches are often time consuming.

In this work, we propose a new method to leverage pre-trained GANs such as StyleGAN~\cite{karras2018style} and BigGAN~\cite{brock2018large} to provide rich and diverse priors for restoration.
This is similar in spirit to the classic notion of dictionary~\cite{yang2010image}, which explicitly constructs a finite image bank.
But unlike the conventional method, we use pre-trained GANs as a latent image bank that is practically infinite, hence can serve as a much stronger prior.
Compared with most GAN inversion methods, which also use pre-trained GANs, our method does not involve image-specific optimization at runtime. Once trained, the model only needs a single forward pass to perform restoration, therefore is more practical for applications that demand fast response.

Conditioning and retrieving from a \textit{GAN-based dictionary} is a new and non-trivial question we need to address in this work. We show that pre-trained GANs can be employed as a latent bank in a succinct \textit{encoder-bank-decoder} architecture.
This novel architecture allows us to lift the burden of simultaneous learning both fidelity and detail generation in a typical encoder-decoder network since the latent bank already captures rich natural image priors.
In addition, we show that it is pivotal to condition the bank by passing the convolutional features from the encoder to achieve high-fidelity results.
We also design a multi-resolution framework for passing features to strengthen information flow from the latent bank to the decoder, further improving the results.
We show the effectiveness of the proposed method in handling images with challenging poses and structures apart from the highly-ill-posed nature of the task. We also demonstrate how the method can be generalized to different categories, \eg~human faces, cats, buildings, by switching different pre-trained GAN latent banks or using more generic priors.

This work is an extension of our earlier conference version~\cite{chan2021glean}. In comparison to the conference version, we have introduced a significant amount of new materials.
\textbf{1)} Through extensive experiments on our original model GLEAN, we find that some modules can be safely removed without sacrificing the performance. Thus, we redesign GLEAN and propose a lightweight version -- \textit{LightGLEAN}. Remarkably, when compared to GLEAN, LightGLEAN consists of only 21\% of parameters while achieving comparable performance, as shown in Fig.~\ref{fig:teaser}(e).
\textbf{2)} In addition to 8$\times$ to 64$\times$ super-resolution, we consider more restoration tasks in this paper. First, we demonstrate the capability of GLEAN on restoring images degraded by complex and unknown real-world degradations.
Second, we show that the natural image prior encapsulated in the latent bank is effective in not only super-resolution but also various restoration tasks such as colorization.
\textbf{3)} We extend GLEAN towards the restoration task of generic images. Different from our conference version that requires different models to restore images of different object classes, we demonstrate the potential of GLEAN on multi-class restorations by employing BigGAN~\cite{brock2018large} as our generative latent bank, which allows class-conditioned image generation. With the use of the multi-class prior, perceptually convincing images of different objects can be restored with a single model.
\vspace{-0.3cm}
% !TEX root = ../submission.tex

\begin{figure*}[!t]
  \begin{center}
    \includegraphics[width=0.95\textwidth]{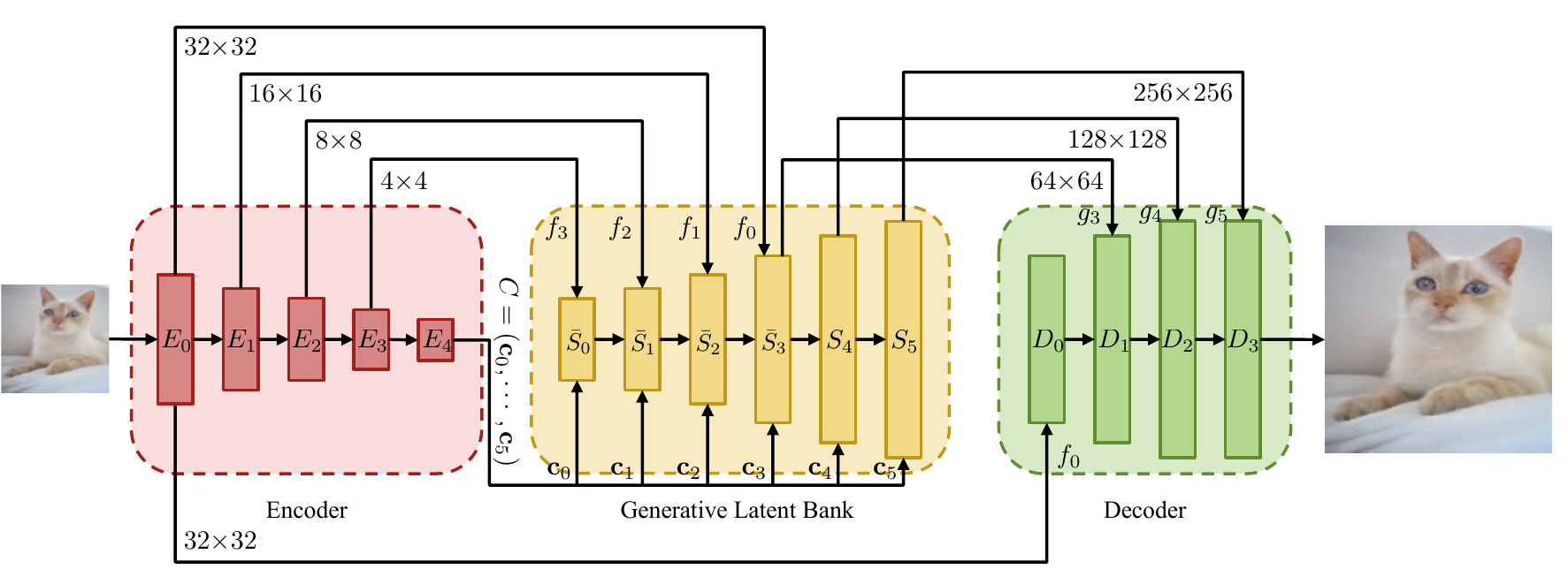}
    \caption{\textbf{Overview of GLEAN}. In addition to the latent vectors $\mathbf{c}_i$, the generator (\ie, the generative latent bank) is also conditioned on the multi-resolution features $f_i$. With a pre-trained GAN capturing the natural image prior, this encoder-bank-decoder design lifts the burden of learning both fidelity and naturalness in the conventional encoder-decoder architecture. $E_i$, $S_i$, and $D_i$ denote the encoder blocks, latent bank blocks, and decoder blocks, respectively. This example corresponds to an input size of $32{\times}32$ and an output size of $256{\times}256$. %Equivalently, $N{=}4$ and $k{=}6$.
    }
    \label{fig:overview}
    \vspace{-0.4cm}
  \end{center}
\end{figure*}

\section{Related Work}
\label{sec:relatedwork}
\noindent{\textbf{Image Super-Resolution.}}
Many existing SR algorithms~\cite{dai2019secondorder,dong2014learning,dong2016image,dong2016accelerating,he2019ode,xu2019towards,zhang2018image,zhou2020cross} directly learn a mapping from low-resolution images to high-resolution images with a pixel-wise constraint (\eg~$\ell_2$ loss).
While these methods achieve remarkable results in terms of PSNR, training solely with pixel-wise constraints often results in perceptually unconvincing outputs with severe over-smoothing artifacts~\cite{ledig2017photo,menon2020pulse}.
To alleviate the problem, GANs~\cite{ledig2017photo,sajjadi2017enhancenet,wang2018esrgan,xu2017learning} are employed to approximate the natural image manifold, yielding more photo-realistic results. For instance, SRGAN~\cite{ledig2017photo} adopts adversarial loss and perceptual loss~\cite{johnson2016perceptual} in addition to $\ell_2$ loss, improving the visual quality of the outputs. However, as the generator needs to learn both fidelity and natural image characteristics, unnatural artifacts could still be observed in the outputs, especially if one trains the generator from scratch.

Recent interests have shifted to large-factor SR beyond the typical upscaling factors ($2{\times}$ or $4{\times}$)~\cite{dahl2017pixel,hyun2020varsr,shang2020perceptual,zhang2019texture}.
Dahl~\etal~\cite{dahl2017pixel} propose a fully probabilistic pixel recursive network for upsampling extremely coarse images with an resolution $8{\times}8$.
RFB-ESRGAN~\cite{shang2020perceptual} builds upon ESRGAN and adopts multi-scale receptive field blocks for $16{\times}$ SR.
VarSR~\cite{hyun2020varsr} achieves $8{\times}$ SR by matching the latent distributions of LR and HR images to recover the missing details.
Zhang~\etal~\cite{zhang2019texture} perform $16{\times}$ reference-based SR on paintings with a non-local matching module and a wavelet texture loss.
To handle even larger magnification factors, one would need to rely on stronger priors. SR methods specializing on large magnification factors are typically dedicated to the human face category as one could exploit the strong structural prior of faces.
Facial priors including facial attributes~\cite{li2019deep}, facial landmarks~\cite{kim2019progressive,ma2020deep}, and identity~\cite{grm2019face} have been studied.
Our work goes beyond previous works by pushing the SR limit to 64${\times}$, a large magnification factor that is challenging due to its highly ill-posed nature, and by generalizing to more categories.

\noindent\textbf{Face Restoration with Generative Prior.}
Several concurrent works adopt generative priors for blind face restoration. Specifically, GFP-GAN~\cite{wang2021towards} employs StyleGAN2 trained on FFHQ dataset to provide facial priors, and incorporate a channel-split SFT~\cite{wang2018recovering} and a facial component loss.
GPEN~\cite{yang2021gan} also makes use of StyleGAN2 as a prior. The noise map in StyleGAN is replaced by a feature map learned from an encoder.
Different from GFP-GAN and GPEN, GLEAN is not confined to face restoration. Instead, we focus on a generic framework that is applicable to a wide range of object categories and tasks.
Moreover, unlike GFP-GAN and GPEN, which use the entire StyleGAN as a prior network, we carefully examine the use of StyleGAN in image restoration and find that using the entire StyleGAN for image restoration is unnecessary. Consequently, we redesign GLEAN by pruning a significant portion of it. Our lightweight model, LightGLEAN, achieves comparable performance while having only 21\% of parameters and 35\% of FLOPs.

\noindent{\textbf{Image Colorization.}}
Existing works~\cite{cheng2015deep,su2020instance,iizuka2016let,zhang2016colorful} generally use pixel-wise loss between the output image and ground-truth image to guide the training of the network. With sufficient training data, learning-based methods are able to achieve promising results. However, it is well-known that pixel-wise losses such as $\ell_2$ loss often lead to images with flat color and reduced perceptual quality. Later works~\cite{deoldify,vitoria2020chromagan,isola2017image} explore the possibility of generative adversarial networks to approximate the natural image manifold for better visual quality. Some studies attempt to employ additional priors such as object class~\cite{su2020instance,vitoria2020chromagan,iizuka2016let} to improve the performance. In this work, we demonstrate a new way of exploiting prior information, particularly the prior captured in generative models for improving color restoration.

\noindent{\textbf{GAN Inversion.}}
Given a degraded image $x$, GAN inversion-based methods~\cite{bau2020semantic,gu2020image,menon2020pulse,pang2020exploiting} in general produce a natural image best approximating $x$ by optimizing $z^* = \argmin\nolimits_{z \in \mathcal{Z}} \mathcal{L}\left(G(z), x\right)$, where $\mathcal{Z}$ is the latent space and $\mathcal{L}(\cdot,\cdot)$ denotes the task-specific objective function.
For instance, PULSE~\cite{menon2020pulse} iteratively optimizes the latent code of StyleGAN~\cite{karras2018style} with a pixel-wise constraint between the input and output.
mGANprior~\cite{gu2020image} optimizes multiple latent codes to increase the expressiveness of the model.
DGP~\cite{pang2020exploiting} further finetunes the generator together with the latent code to reduce the gap between the distributions of the training and testing images.
A common issue with GAN inversion is that important spatial information may not be faithfully retained due the low-dimensionality of the latent code. Thus, these methods often generate undesirable results that do not resemble the ground-truth.
Different from GAN inversion, GLEAN conditions the pre-trained generator on both the latent codes and multi-resolution convolutional features, providing additional spatial guidance for restoration. In addition, GLEAN does not require iterative optimization during inference.

\vspace{-0.15cm}
% !TEX root = ../submission.tex

\section{GLEAN}
\label{sec:method}
A GAN model that is trained on large-scale natural images captures rich texture, color, and shape priors.
Previous studies~\cite{bau2020semantic,gu2020image,menon2020pulse,pang2020exploiting} have shown that such priors can be harvested through GAN inversion to benefit various image restoration tasks. Nonetheless, methods for exploiting these priors without the costly optimization during inversion remain underexplored.

In this study, we devise GLEAN within a novel \textit{encoder-bank-decoder} architecture, allowing one to exploit the generative priors with just a single forward pass. An overview of the architecture is depicted in Fig.~\ref{fig:overview}.
Given a degraded image, GLEAN applies an encoder to extract latent vectors and multi-resolution convolutional features, which capture important high-level cues as well as spatial structure of the LR image.
Such cues are used to condition the latent bank, further producing another set of multi-resolution features for the decoder.
Finally, the decoder generates the final output by integrating the features from both the encoder and the latent bank.
In this work, we adopt the state-of-the-art GAN architectures as the generative latent bank, such as StyleGAN~\cite{karras2018style,karras2019analyzing} and BigGAN~\cite{brock2018large}, while the specific choice is flexible, depending on different applications.

\subsection{Encoder}
To generate the latent vectors, we first use an RRDBNet~\cite{wang2018esrgan} (denoted as $E_0$) to extract features $f_0$ from the input LR image.
Then, we gradually reduce the resolution of the features by: % using $N$ strided convolutions:
\begin{equation}
	f_i = E_i(f_{i-1}),\quad i\in\{1, \cdots, N\},
\end{equation}
where $E_i$, ${i\in\{1, \cdots, N\}}$, denotes a stack of a stride-2 convolution and a stride-1 convolution. Finally, a convolution and a fully-connected layer are used to generate the latent vectors:
\begin{equation}
	C = E_{N + 1}(f_{N}),
\end{equation}
where $C$ is a matrix whose columns represent the latent vectors for the generative latent bank.

The latent vectors in $C$ capture a compressed representation of the images, providing the generative latent bank with high-level information.
To further capture the local structures of the LR image and to provide additional guidance for structure restoration, we also feed multi-resolution convolutional features $\{f_i\}$ into the latent bank.

\subsection{Generative Latent Bank}
Given the convolutional features $\{f_i\}$ and the latent vectors $C$, we leverage a pre-trained generator as a latent bank to provide priors for texture and detail generation.
As GAN is originally designed for image generation tasks, it cannot be directly integrated into the proposed encoder-bank-decoder framework.
In this work, we adapt the GAN architecture (\eg~StyleGAN and BigGAN) to our framework by making three modifications:

\textbf{1)} Instead of taking one single latent vector as the input, each block of the generator takes a different latent vector to improve expressiveness. More specifically, we have $C{=} (\mathbf{c}_0, \cdots, \mathbf{c}_{k-1})$ for $k$ blocks, where each $\mathbf{c}_i$ corresponds to one latent vector. We find that this modification leads to outputs with fewer artifacts. This modification is also seen in previous works~\cite{gu2020image,xu2020generative,zhu2020in}.

\textbf{2)} To allow conditioning on the additional features from the encoder, we use an additional convolution in each style block for feature fusion for features whose resolution is smaller than or equal to the input resolution:
\begin{equation}
	g_i =
	\begin{cases}
		\bar{S}_0(\mathbf{c}_0, f_{N}),              & \text{if }i = 0,      \\
		\bar{S}_i(\mathbf{c}_{i}, g_{i-1}, f_{N-i}), & \text{if }0< i\leq N, \\
		S_i(\mathbf{c}_{i}, g_{i-1}),                & \text{otherwise},     \\
	\end{cases}
\end{equation}
where $S_i$ and $\bar{S}_i$ denote the original style block and the augmented style block with an additional convolution, respectively. $g_i$ corresponds to the output feature of the $i$-th style block.

\textbf{3)} Instead of directly generating outputs from the generator, we output the features $\{g_i\}$ and pass them to the decoder to better fuse the features from the latent bank and the encoder.

\noindent{\textbf{Advantages.}} The use of generative latent bank is reminiscent of the task of reference-based restoration, where external HR information, such as single reference image~\cite{shim2020robust,xie2020feature,yang2020learning,zhang2020copy,zhang2019image,zheng2018crossnet}, multiple reference images~\cite{li2020blind,li2020enhanced,yan2020towards} and learnable dictionary~\cite{li2020blind}, is used.
While the external HR information leads to marked improvements, the performance is sensitive to the similarity between the inputs and references. This sensitivity may eventually lead to degraded results when the reference images/components are not well selected.
Moreover, the size and diversity of those imagery dictionaries are limited by the selected components, impeding the generalization to diverse scenes in practice.
In addition, computationally-intensive global matching~\cite{zhang2019image} or component detection/selection~\cite{li2020blind} is often required to aggregate appropriate information from the references, hindering the applications to scenarios with tight computational constraints.
Instead of constructing an imagery dictionary, GLEAN adopts a \textit{GAN-based} dictionary conditioned on a pre-trained GAN. Our dictionary does not depend on any specific components or images. Instead, it captures the distribution of the images and has potentially unlimited size and diversity.
Furthermore, GLEAN is computationally efficient without requiring global matching and the reference images/components selection.

\subsection{Decoder}
GLEAN uses an additional decoder with progressive fusion to integrate the features from the encoder and the latent bank to generate output image.
It takes the RRDBNet features as inputs and progressively fuses the features with the multi-resolution features from the latent bank:
\begin{equation}
	d_i =
	\begin{cases}
		D_0(f_0)                    & \text{if } i = 0, \\
		D_i(d_{i-1}, g_{N - 1 + i}) & \text{otherwise},
	\end{cases}
\end{equation}
where $D_i$ and $d_i$ denote a $3{\times}3$ convolution and its output, respectively. Each convolution is followed by a pixel-shuffle~\cite{shi2016real} layer except the final output layer.
With the skip-connection between the encoder and decoder, the information captured by the encoder can be reinforced, and hence the latent bank could better focus on the texture and detail generation.

\subsection{Training}
Similar to existing works~\cite{ledig2017photo,wang2018recovering,wang2018esrgan}, we adopt the standard MSE loss, perceptual loss~\cite{johnson2016perceptual}, and adversarial loss for training. MSE loss is used to guide the fidelity of the output images:
\begin{equation}
	\mathcal{L}_{mse} = \dfrac{1}{N}||\hat{y} - y||_2^2,
\end{equation}
where $N$, $\hat{y}$, and $y$ denote the number of pixels, the output image, and the ground-truth image, respectively. We further incorporate perceptual loss~\cite{johnson2016perceptual} and adversarial loss~\cite{goodfellow2014generative} to improve the perceptual quality:
\begin{align}
	 & \mathcal{L}_{percep} = \dfrac{1}{N}||V(\hat{y}) - V(y)||_2^2,   \\
	 & \mathcal{L}_{gen} = \log \left(1 - \mathcal{D}(\hat{y})\right),
\end{align}
where $V(\cdot)$ denotes the feature embedding space of the VGG16~\cite{simonyan2014very} network, and $\mathcal{D}$ corresponds to the StyleGAN or BigGAN discriminator. The resulting objective function is a weighted mean of the three losses:
\begin{equation}
	\mathcal{L}_g = \mathcal{L}_{mse} + \alpha_{percep}{\cdot}\mathcal{L}_{percep} + \alpha_{gen}{\cdot}\mathcal{L}_{gen}.
\end{equation}
In all our experiments, we set $\alpha_{percep}{=}\alpha_{gen}{=}10^{-2}$. For the discriminator, we minimize
\begin{equation}
	\mathcal{L}_{d} = - ( \log \left(1 - \mathcal{D}(\hat{y})\right) + \log \mathcal{D}(y) ).
\end{equation}

To exploit the generative prior, we keep the weights of the latent bank fixed throughout training.
This is because the latent bank may eventually be biased to the training distribution and can potentially harm the model generalizatbility.
It is worth emphasizing that despite GLEAN being trained with similar objectives as in existing works (\eg~ESRGAN), the main difference to these methods is that GLEAN leverages a pre-trained generator to directly incorporate the priors into the network, further improving the output quality. We show that the improvement is not due to additional parameters in the generator by comparing GLEAN with a larger version of ESRGAN, named ESRGAN$^+$.

\begin{figure}[!t]
	\begin{center}
		\includegraphics[width=0.49\textwidth]{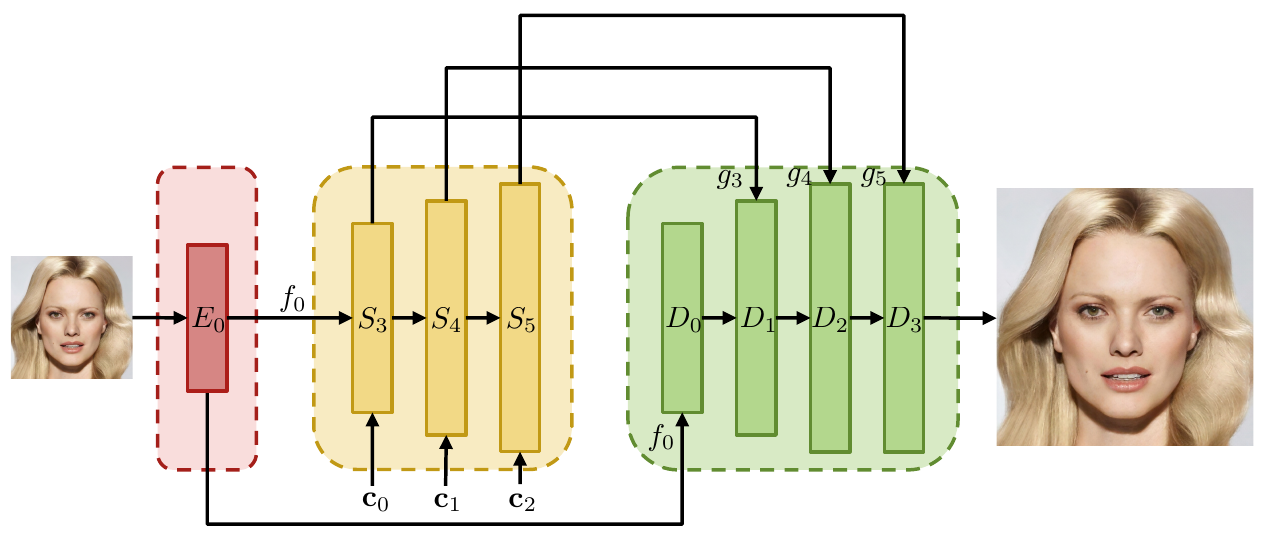}
		\caption{\textbf{Overview of LightGLEAN}. Unlike GLEAN, which generates features with resolution down to $4{\times}4$, the latent bank in LightGLEAN directly conditions on the RRDB feature $f_0$, bypassing the style blocks that corresponds to the coarse resolutions. In addition, a fixed latent code $\mathbf{c}$ is used for all style blocks. In this design, LightGLEAN can be devised with much fewer learnable parameters.
		}
		\label{fig:overview_light}
	\end{center}
\end{figure}
\begin{table}[!t]
	\vspace{-0.5cm}
	\caption{\textbf{Complexity comparison between LightGLEAN and GLEAN.} LightGLEAN contains 79\% fewer parameters. Comparison is performed on the models for $64{\times}64$ input and $1024{\times}1024$ output.}
	\label{tab:quan_light}
	\begin{center}
		\tabcolsep=0.1cm
		\vspace{-0.3cm}
		\scalebox{1}{
			\begin{tabular}{l|c|c|c}
				          & GLEAN  & LightGLEAN & \% Reduction \\\hline
				Encoder   & 137.3M & 23.7M      & 82.7\%       \\
				Generator & 30.4M  & 10.9M      & 63.9\%       \\
				Decoder   & 7.9M   & 1.7M       & 78.6\%       \\\hline
				Params    & 175.6M & 36.3M      & 79.3\%       \\
				FLOPs     & 277.5G & 98.24G     & 64.6\%       \\
			\end{tabular}}
	\end{center}
\end{table}
\begin{figure}[!t]
	\begin{center}
		\vspace{-0.4cm}
		\includegraphics[width=0.49\textwidth]{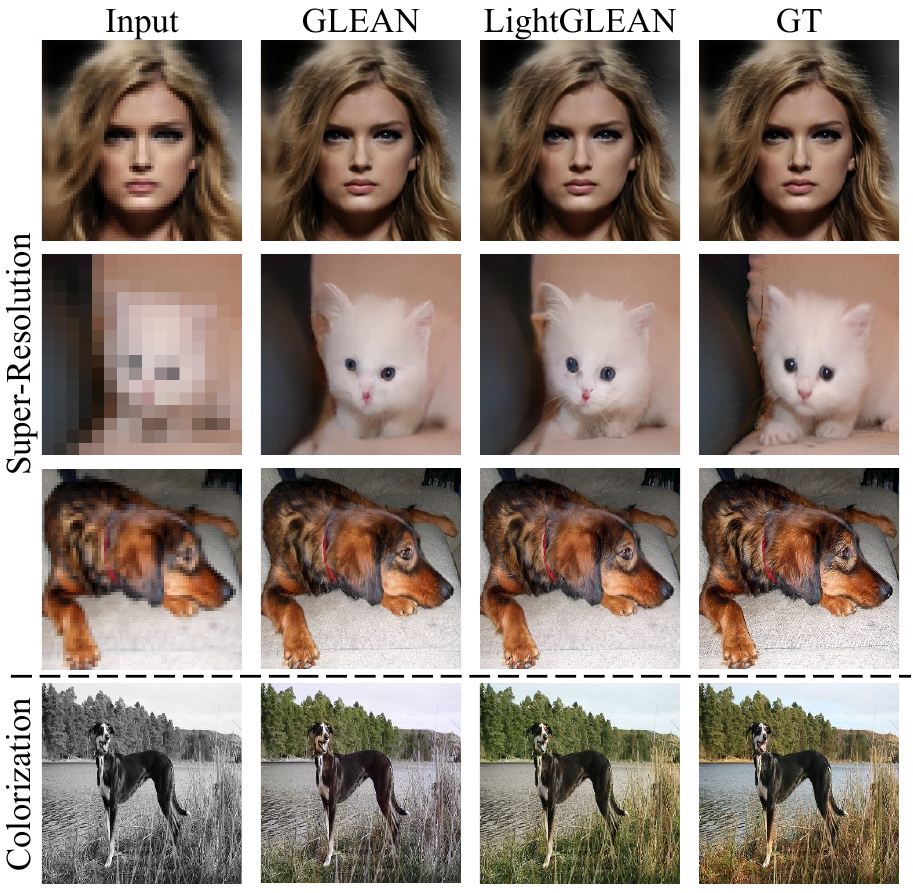}
		\vspace{-0.7cm}
		\caption{\textbf{Qualitative comparison between GLEAN and LightGLEAN.} Despite being more lightweight than GLEAN, LightGLEAN provides outputs that are comparable to GLEAN. \textbf{(Zoom in for best view)}}
		\label{fig:lightglean_quali}
		\vspace{-0.6cm}
	\end{center}
\end{figure}
\section{LightGLEAN}
Although it achieves remarkable performance, GLEAN has a large model size. To address this issue, we conduct an in-depth analysis of the design of GLEAN, such that nonessential modules can be identified and pruned.

As shown in Fig.~\ref{fig:overview}, GLEAN adopts a multi-resolution structure, and the number of feature channels increases when resolution decreases.
As a result, the encoder incurs a significant number of parameters in modules corresponding to the coarse features.
While it is a common approach to synthesize high-quality images from a coarse resolution such as a $4{\times}4$ input (\eg~StyleGAN and BigGAN), we find that the features with a resolution smaller than that of the input image are less important in the task of image restoration, since the feature $f_0$ extracted from the input image has already provided rich spatial and structural information for the subsequent process in the generative latent bank.

We propose the following two strategies to simplify the structure of GLEAN, and the resulting lightweight model, LightGLEAN, is shown in Fig.~\ref{fig:overview_light}.

\textbf{1)} We remove the coarse feature connections between the encoder and the generator.
Specifically, instead of generating coarse features and performing feature fusion, only $E_0$ is kept in the encoder of LightGLEAN. The encoder outputs only the RRDB feature $f_0$, and only $f_0$ is sent to the generator. For the generator, let $i_0$ be the style block index corresponding to the input resolution, the modules for the coarse resolutions (\ie, $\bar{S}_i, i{\in}\{i{=}1,{\cdots}, i_0{-}1\}$) are bypassed, and the feature fusion modules are removed. The encoder feature $f_0$ is directly used as an input to the style block for the finer resolutions. Symbolically, our latent bank is modified as follows:
\begin{equation}
	g_i =
	\begin{cases}
		S_i(\mathbf{c}_i, f_0),     & \text{if }i = i_0, \\
		S_i(\mathbf{c}_i, g_{i-1}), & \text{if }i > i_0.
	\end{cases}
\end{equation}

\textbf{2)} In LightGLEAN, the latent codes $\mathbf{c}_i$ are no longer learned from the encoder. Instead, they are casted as learnable parameters, and the same set of latent codes is applied for all input images. It further reduces the number of parameters as the linear layers are omitted.

By removing the coarse resolution modules that contribute to a significant portion of the parameters, a lightweight architecture is devised. Note that LightGLEAN can be further pruned by reducing the number of RRDBs~\cite{wang2018esrgan} without significant performance drop\footnote{In the task of $16{\times}$ face super-resolution, the PSNR merely drops by 0.02 dB when reducing the number of RRDBs from 23 to 10.}. Such exploration is left as our future work.

As shown in Table~\ref{tab:quan_light}, LightGLEAN has only 21\% of parameters when compared to GLEAN. Notably, the encoder of GLEAN consists of 137.3M parameters, making it hard to deploy in practice. In contrast, with our careful pruning, the encoder of LightGLEAN contains only 23.7M parameters, which is 82.7\% fewer parameters than that of GLEAN. LightGLEAN achieves a comparable performance to GLEAN with 79\% reduction of parameters and 65\% reduction of FLOPs.
The examples in Fig.~\ref{fig:lightglean_quali} show that the output quality of LightGLEAN and GLEAN are comparable.
More quantitative comparisons are discussed in the next section.

% !TEX root = ../submission.tex
\begin{figure*}[!t]
	\begin{center}
		\includegraphics[width=0.99\textwidth]{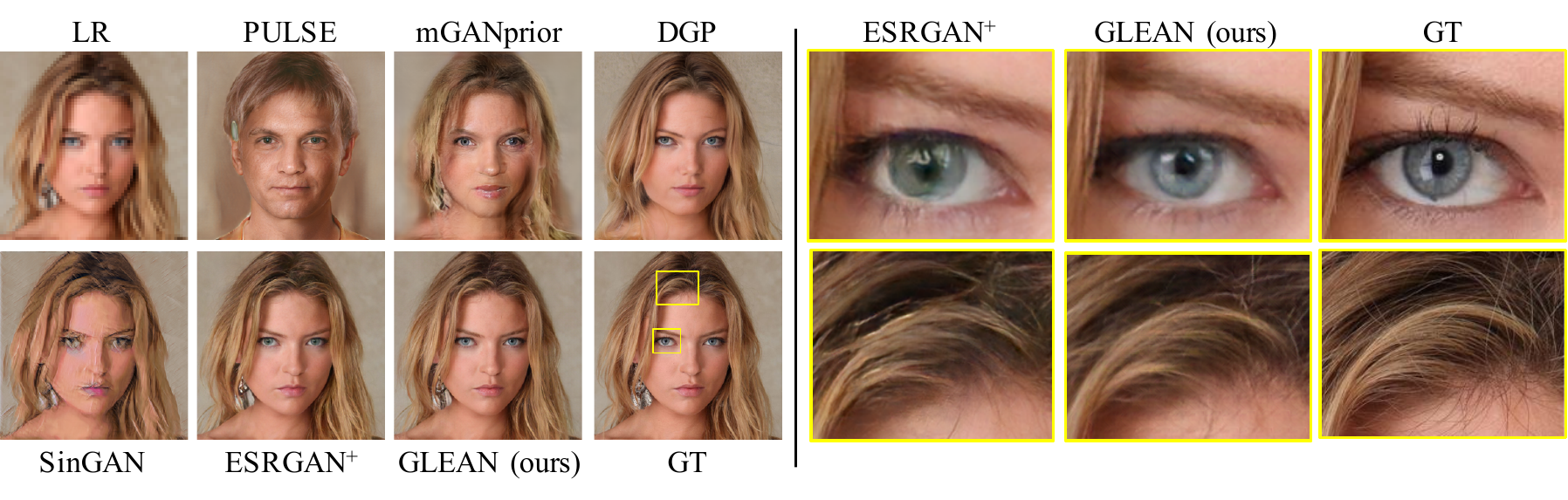}
		% \vskip -0.25cm
		\caption{\textbf{Comparisons on 16${\times}$ SR on CelebA-HQ~\cite{karras2018progressive}.} Only GLEAN is able to maintain high fidelity while synthesizing realistic textures and details: GAN inversion methods fail to preserve the identity, and adversarial loss methods struggle to synthesize fine details. ESRGAN$^{+}$ denotes a larger version with similar runtime to GLEAN. \textbf{(Zoom in for best view)}}
		\label{fig:sr}
		\vspace{-0.5cm}
	\end{center}
\end{figure*}
\section{Experiments}
\label{sec:exp}
\subsection{Training Details}
\label{sec:training}
We adopt pre-trained StyleGAN\footnote{GenForce: https://github.com/genforce/genforce}~\cite{karras2018style}, StyleGAN2\footnote{BasicSR: https://github.com/xinntao/BasicSR}$^{,}$\footnote{MMEditing~\cite{mmedit}: https://github.com/open-mmlab/mmediting}~\cite{karras2019analyzing}, or BigGAN\footnote{https://github.com/ajbrock/BigGAN-PyTorch}~\cite{brock2018large} as our latent bank using the publicly available models and codes.

We train and test GLEAN on various datasets. The training and test datasets used in our experiments are summarized in Table~\ref{tab:dataset}. For fair comparisons, we train the baselines on the same datasets as our model. The test set is strictly exclusive from training.
Since StyleGAN and BigGAN produce images with a fixed size, we resize the images in the datasets for our experiments. The specific degradations are described in each section.
\begin{table}[!t]
	\caption{\textbf{Datasets used in our experiments.}}
	\label{tab:dataset}
	\begin{center}
		\tabcolsep=0.15cm
		\vspace{-0.5cm}
		\scalebox{0.95}{
			\begin{tabular}{l|c|c}
				                               & \textbf{Train}                         & \textbf{Test}                                          \\\hline
				\textbf{Human faces (Bicubic)} & FFHQ~\cite{karras2018style}            & CelebA-HQ~\cite{karras2018progressive}                 \\
				\textbf{Human faces (Blind)}   & FFHQ~\cite{karras2018style}            & LFW~\cite{huang2008labeled}, CelebA~\cite{liu2015deep} \\
				\textbf{Cats}                  & LSUN-train~\cite{yu2015lsun}           & CAT~\cite{zhang2008cat}                                \\
				\textbf{Cars}                  & LSUN-train~\cite{yu2015lsun}           & Cars~\cite{krause2013object}                           \\
				\textbf{Bedrooms}              & LSUN-train~\cite{yu2015lsun}           & LSUN-validate~\cite{yu2015lsun}                        \\
				\textbf{Towers}                & LSUN-train~\cite{yu2015lsun}           & LSUN-validate~\cite{yu2015lsun}                        \\
				\textbf{Multi-class}           & ImageNet-train~\cite{deng2009imagenet} & ImageNet-val~\cite{deng2009imagenet}                   \\
			\end{tabular}}
	\end{center}
\end{table}

We adopt Cosine Annealing scheme~\cite{loshchilov2016sgdr} and Adam optimizer~\cite{kingma2014adam} in training. The number of iterations is 300K and the initial learning rate is $10^{-4}$. The batch size is 8 for human faces, 16 for other class-specific training, and 32 for ImageNet training.

\begin{figure}[!t]
	\begin{center}
		\includegraphics[width=0.49\textwidth]{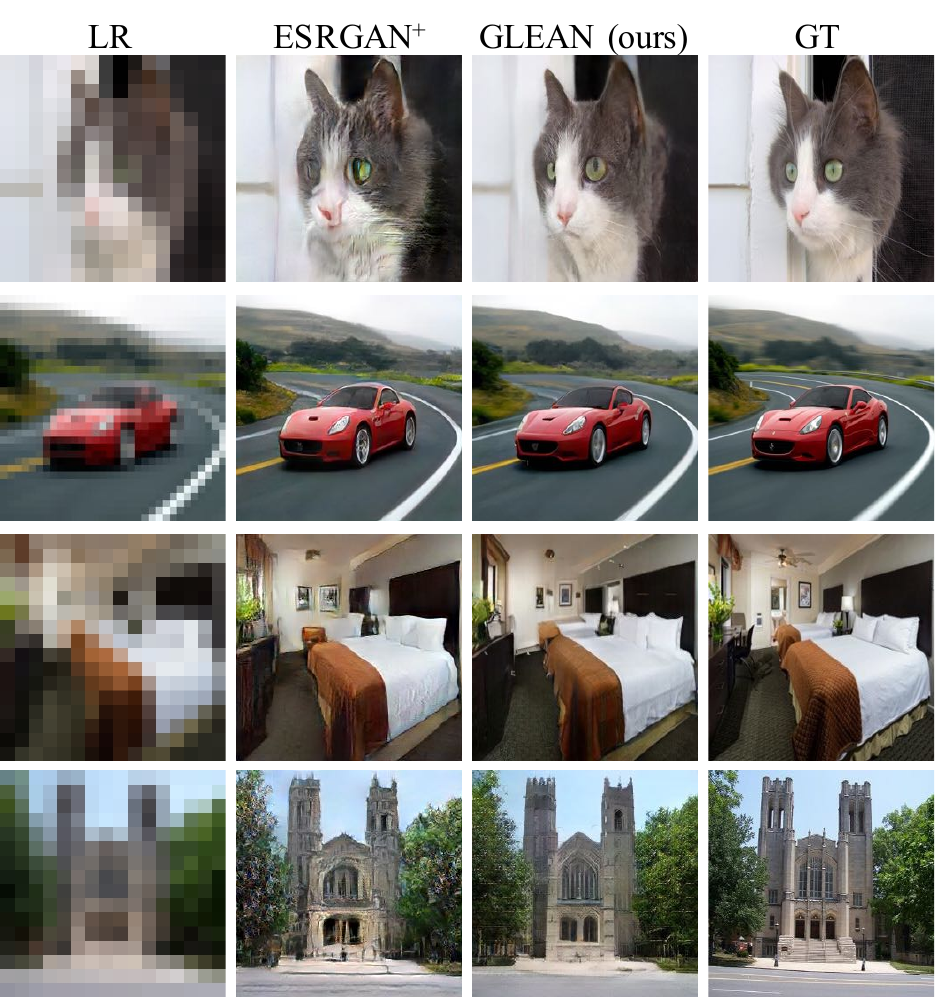}
		\caption{\textbf{Results of 16${\times}$ SR on other categories.} GLEAN can be applied to various categories by switching between StyleGANs trained on different categories. \textbf{(Zoom in for best view)}}
		\label{fig:sr_others}
	\end{center}
	\vspace{-0.5cm}
\end{figure}
\begin{figure}[!t]
	\begin{center}
		\includegraphics[width=0.49\textwidth]{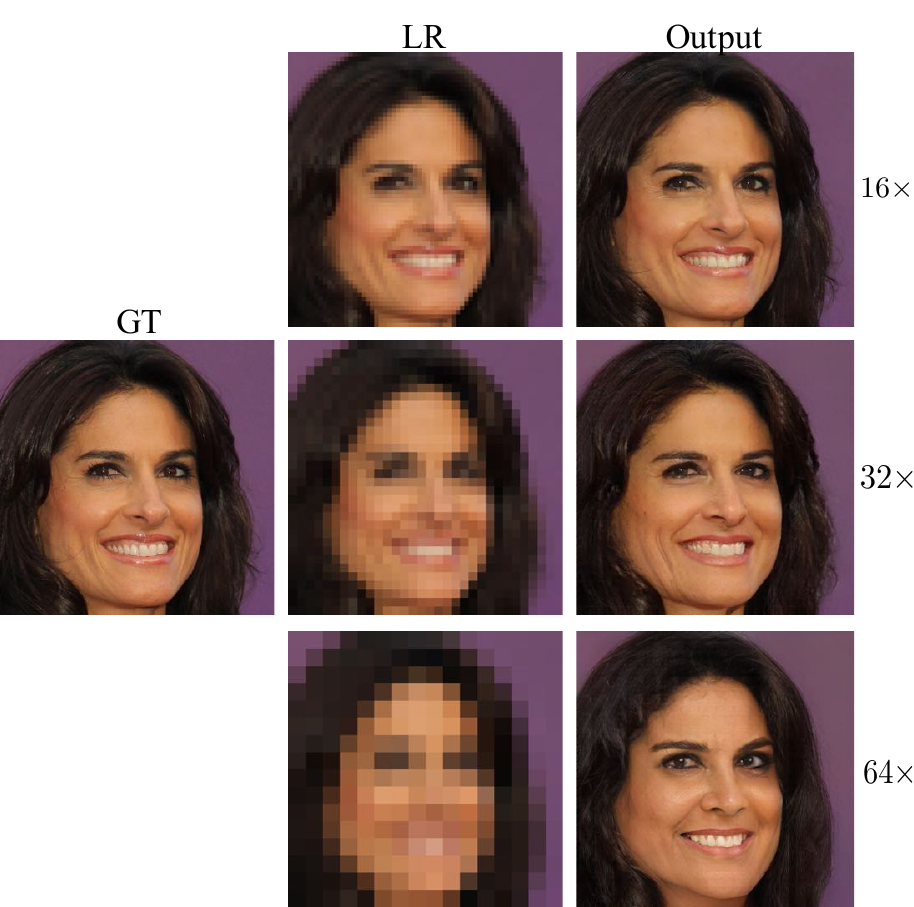}
		% \caption{With the structural information provided by the encoder and the guidance in high-resolution image space, GLEAN is able to generate realistic images highly similar to the ground-truth, even in a large scale of $64{\times}$. \textbf{(Zoom in for best view)}}
		\caption{\textbf{Results on larger scale factors.} GLEAN reconstructs realistic images highly similar to the GT for up to $64{\times}$ upscaling factor. \textbf{(Zoom in for best view)}}
		\label{fig:sr_scale}
	\end{center}
\end{figure}

\subsection{Class-Specific Super-Resolution}
In this section, we assume the downsampling kernel is known, and we synthesize training and test data using the same degradation. Specifically, bicubic downsampling is used to synthesize LR-HR pairs.

The qualitative comparison on $16{\times}$ SR is shown in Fig.~\ref{fig:sr}. Since the GAN inversion methods are only guided by low-dimensional vectors and constraints in LR space, they are unable to maintain a good fidelity of the outputs. In particular, PULSE~\cite{menon2020pulse} and mGANprior~\cite{gu2020image} fail to restore a face image with the same identity. In addition, artifacts are observed in their outputs.
Through finetuning the generator during optimization, the result of DGP~\cite{pang2020exploiting} demonstrates significant improvements in both quality and fidelity. However, a slight difference between the identities of the output and ground-truth is still observed. For example, the eyes and lips show noticeable differences.

Methods trained with adversarial loss (SinGAN~\cite{shaham2019singan}, ESRGAN$^+$\footnote{A larger version of ESRGAN with similar runtime to GLEAN.}~\cite{wang2018esrgan}) can preserve the local structures, but fail in synthesizing convincing textures and details.
Specifically, SinGAN fails to capture the natural image style, producing a painting-like image. Although ESRGAN$^+$ is capable of generating a realistic image, it struggles to synthesize fine details and introduces unnatural artifacts in detailed regions. It is worth emphasizing that although ESRGAN$^+$ achieves competitive results on human faces, its perceptual quality on other categories such as \textit{cats} and \textit{cars} are less promising (see Fig.~\ref{fig:teaser} and Fig.~\ref{fig:sr_others}).
With the latent bank providing natural image priors, GLEAN succeeds in both fidelity and naturalness.
For example, when compared to ESRGAN$^+$, GLEAN reconstructs eyes with better shape and details.
We further extend our method to larger scale factors in Fig.~\ref{fig:sr_scale}. GLEAN successfully generates realistic images resembling the ground truth up to $64{\times}$ upscaling.
\begin{figure}[!t]
	\begin{center}
		\includegraphics[width=0.45\textwidth]{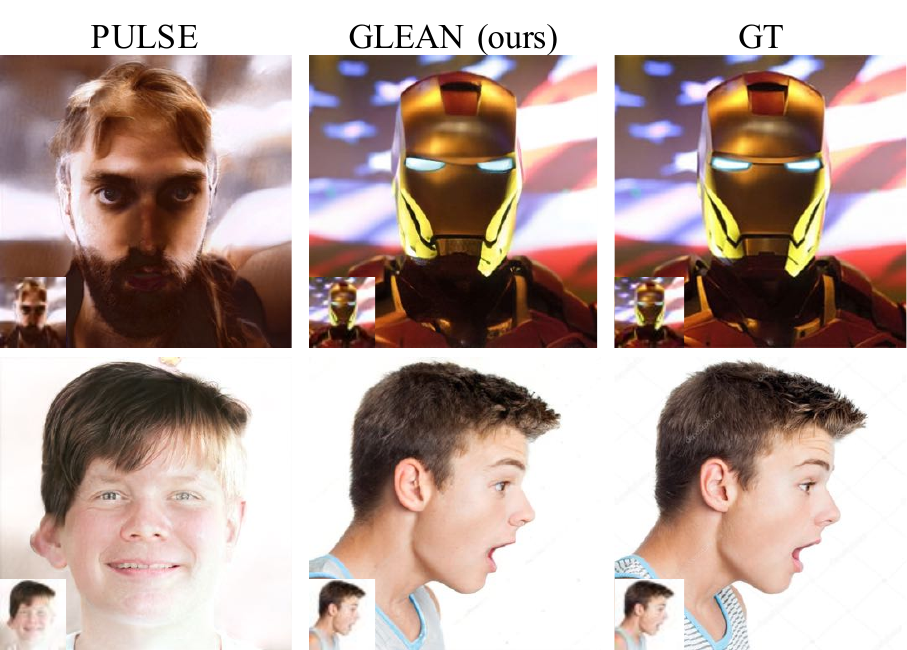}
		\caption{\textbf{Outputs with diverse poses and contents.} Despite GLEAN being trained with aligned human faces, it is able to reconstruct faithful images for non-aligned and non-human faces. PULSE approximates the GT in low resolution \textit{(inlet image at the bottom left corner)}, but its outputs are significantly different from the GT when viewed in high resolution.}
		\label{fig:sr_pose}
	\end{center}
	\vspace{-0.3cm}
\end{figure}

\begin{table}[!t]
	\caption{\textbf{Cosine similarity of ArcFace features~\cite{deng2019arcface} for 16${\times}$ SR.} GLEAN and LightGLEAN achieve a higher similarity than baselines. \textbf{Bolded} texts represent the best performance.}
	\label{tab:arcface}
	\begin{center}
		\tabcolsep=0.15cm
		\vspace{-0.5cm}
		\scalebox{0.89 }{
			\begin{tabular}{l|c|c|c|c}
				           & Bicubic                        & PULSE~\cite{menon2020pulse}      & mGANprior~\cite{gu2020image} & DGP~\cite{pang2020exploiting} \\\hline
				Similarity & 0.8939                         & 0.4047                           & 0.5526                       & 0.7341                        \\ \hline\hline
				           & SinGAN~\cite{shaham2019singan} & ESRGAN$^+$~\cite{wang2018esrgan} & \textbf{GLEAN}               & \textbf{LightGLEAN}           \\\hline
				Similarity & 0.7718                         & 0.9599                           & \textbf{0.9678}              & 0.9607
			\end{tabular}}
	\end{center}
\end{table}
\begin{table}[!t]
	\caption{\textbf{Quantitative (PSNR/LPIPS) comparison on 16${\times}$ SR.} GLEAN outperforms other methods in most categories. \textbf{Bolded} texts represent the best performance.}
	\label{tab:quan}
	\begin{center}
		\tabcolsep=0.05cm
		\vspace{-0.5cm}
		\scalebox{0.85}{
			\begin{tabular}{l|c|c|c|c|c}
				        & mGANprior~\cite{gu2020image} & PULSE~\cite{menon2020pulse} & ESRGAN$^+$~\cite{wang2018esrgan} & \textbf{GLEAN}                 & \textbf{LightGLEAN}   \\\hline
				Face    & 23.66/0.4661                 & 21.83/0.4600                & 26.76/0.2787                     & 26.84/\textbf{0.2681}          & \textbf{26.85}/0.2784 \\
				Cat     & 17.01/0.5556                 & 19.78/0.5241                & 19.99/0.3482                     & \textbf{20.92}/\textbf{0.3215} & 20.83/0.3243          \\
				Car     & 14.53/0.7228                 & 16.30/0.6491                & 19.42/0.3006                     & \textbf{19.74}/\textbf{0.2830} & 19.46/0.2887          \\
				Bedroom & 16.38/0.5439                 & 12.97/0.7131                & \textbf{19.47}/\textbf{0.3291}   & 19.44/0.3310                   & 19.37/0.3345          \\
				Tower   & 15.96/0.4870                 & 13.62/0.7066                & 17.86/0.3132                     & \textbf{18.41}/\textbf{0.2850} & 18.28/0.2933          \\
			\end{tabular}}
		\vspace{-0.5cm}
	\end{center}
\end{table}
\begin{figure*}[!t]
	\begin{center}
		\includegraphics[width=0.99\textwidth]{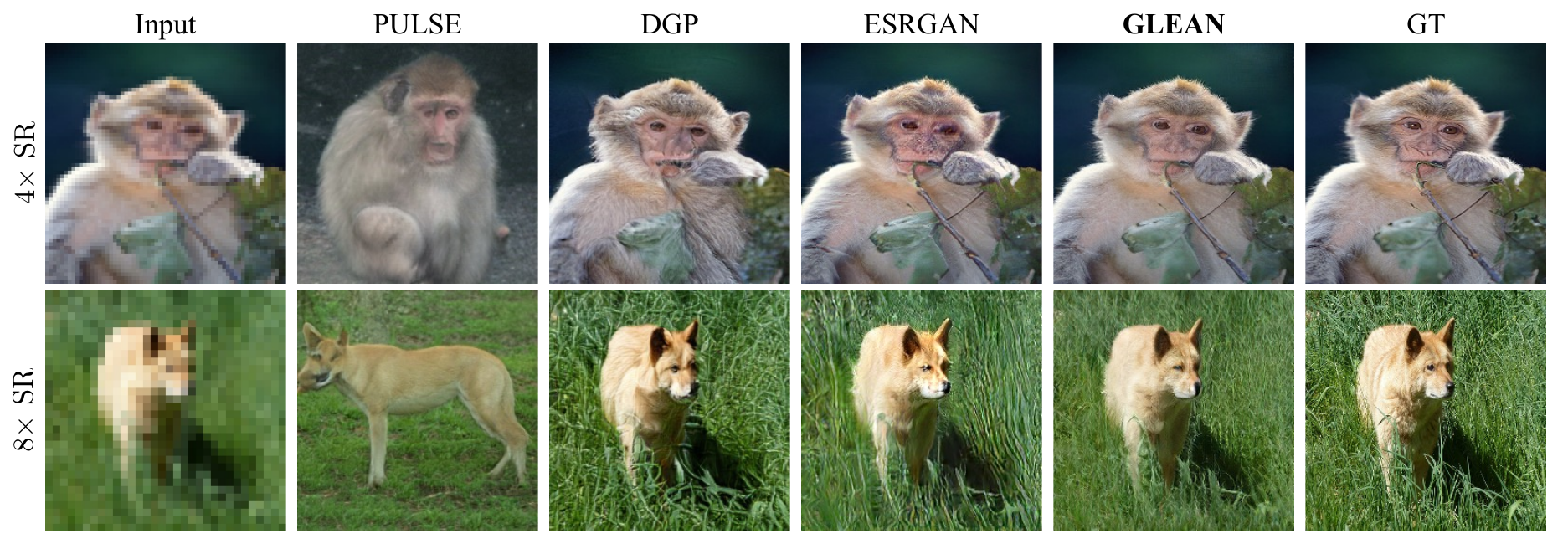}
		\vspace{-0.3cm}
		\caption{\textbf{Results of Super-Resolution using BigGAN.} By employing the multi-class prior encapulated in BigGAN~\cite{brock2018large}, GLEAN can be applied to multiple classes using a single model. GLEAN outperforms existing works in terms of both fidelity and quality. \textbf{(Zoom in for best view)}}
		\label{fig:multi_sr}
		\vspace{-0.3cm}
	\end{center}
\end{figure*}
\begin{figure*}[!t]
	\begin{center}
		\includegraphics[width=0.95\textwidth]{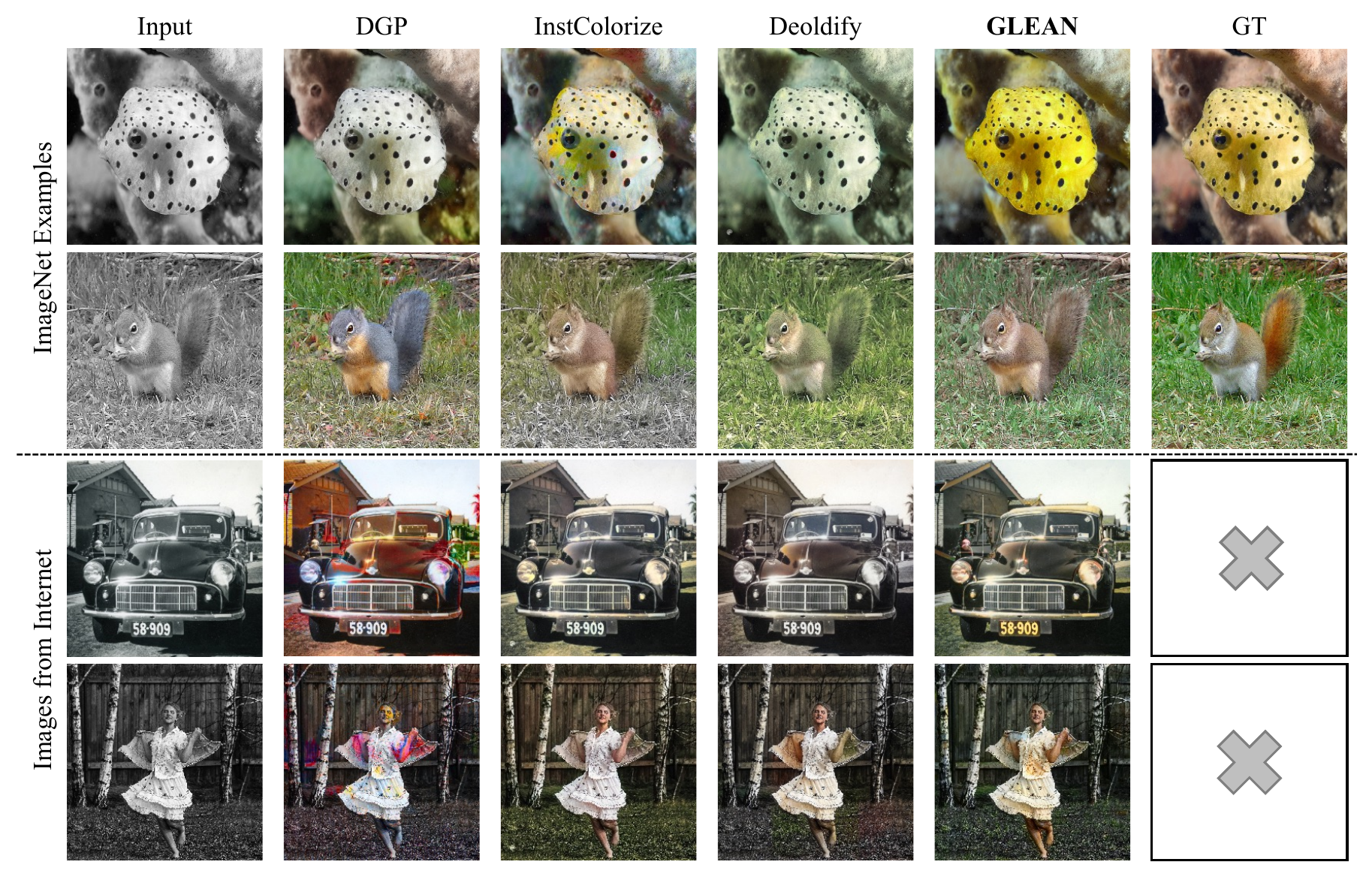}
		\vspace{-0.3cm}
		\caption{\textbf{Results of Image colorization.} In addition to super-resolution, GLEAN can be extended to other restoration tasks such as colorization. By employing the generative priors in BigGAN, GLEAN tends to produce natural color when compared to existing works. GLEAN is also applicable to real-world old photos.}
		\label{fig:multi_color}
		\vspace{-0.5cm}
	\end{center}
\end{figure*}
\begin{figure*}[!t]
	\begin{center}
		\includegraphics[width=0.95\textwidth]{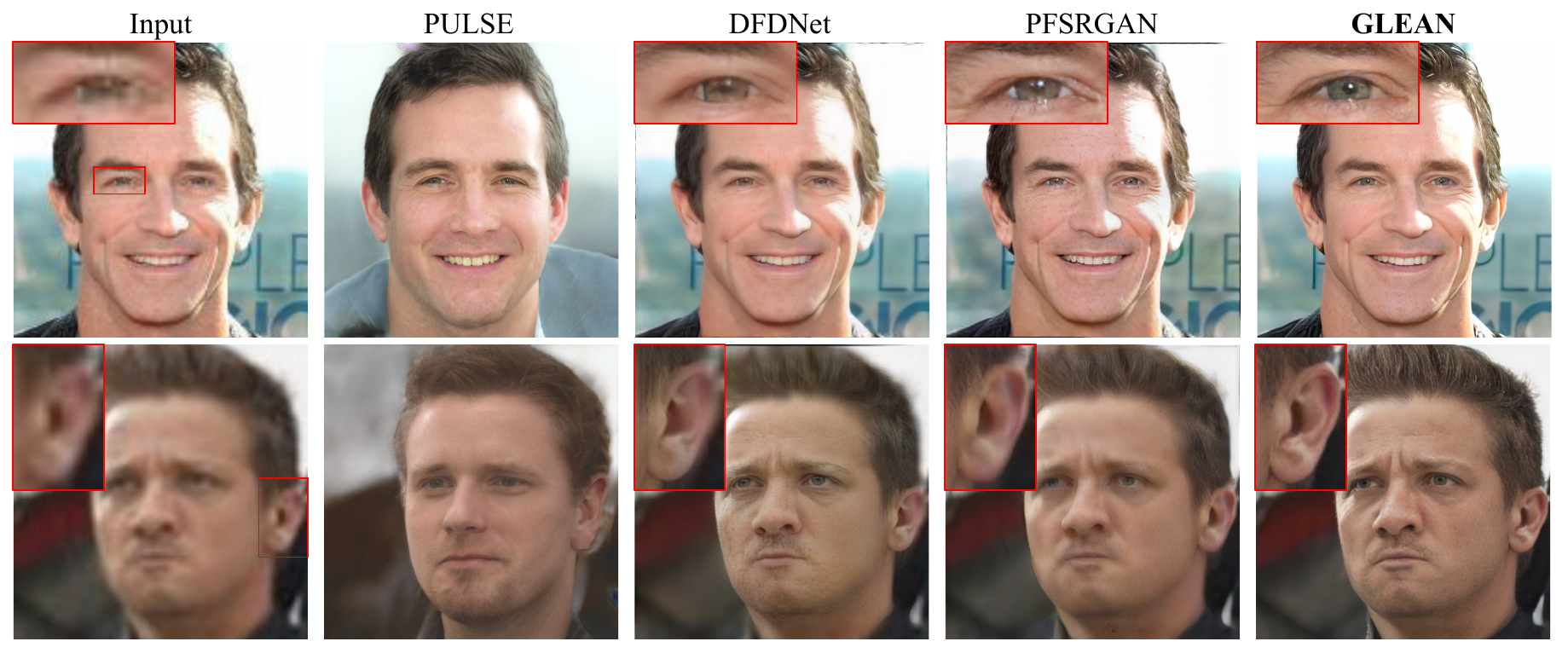}
		\vspace{-0.3cm}
		\caption{\textbf{Results of real-world image restoration.} GLEAN can be extended to unknown degradations by applying various degradations during training. With the natural image prior encapsulated in our latent bank, GLEAN is able to produce results with high quality and natural textures, whereas existing methods produce outputs either  with low fidelity, noticeable artifacts, or blurry details. \textbf{(Zoom in for best view)}}
		\label{fig:blind}
		\vspace{-0.5cm}
	\end{center}
\end{figure*}
\begin{figure*}[!t]
	\begin{center}
		\includegraphics[width=0.95\textwidth]{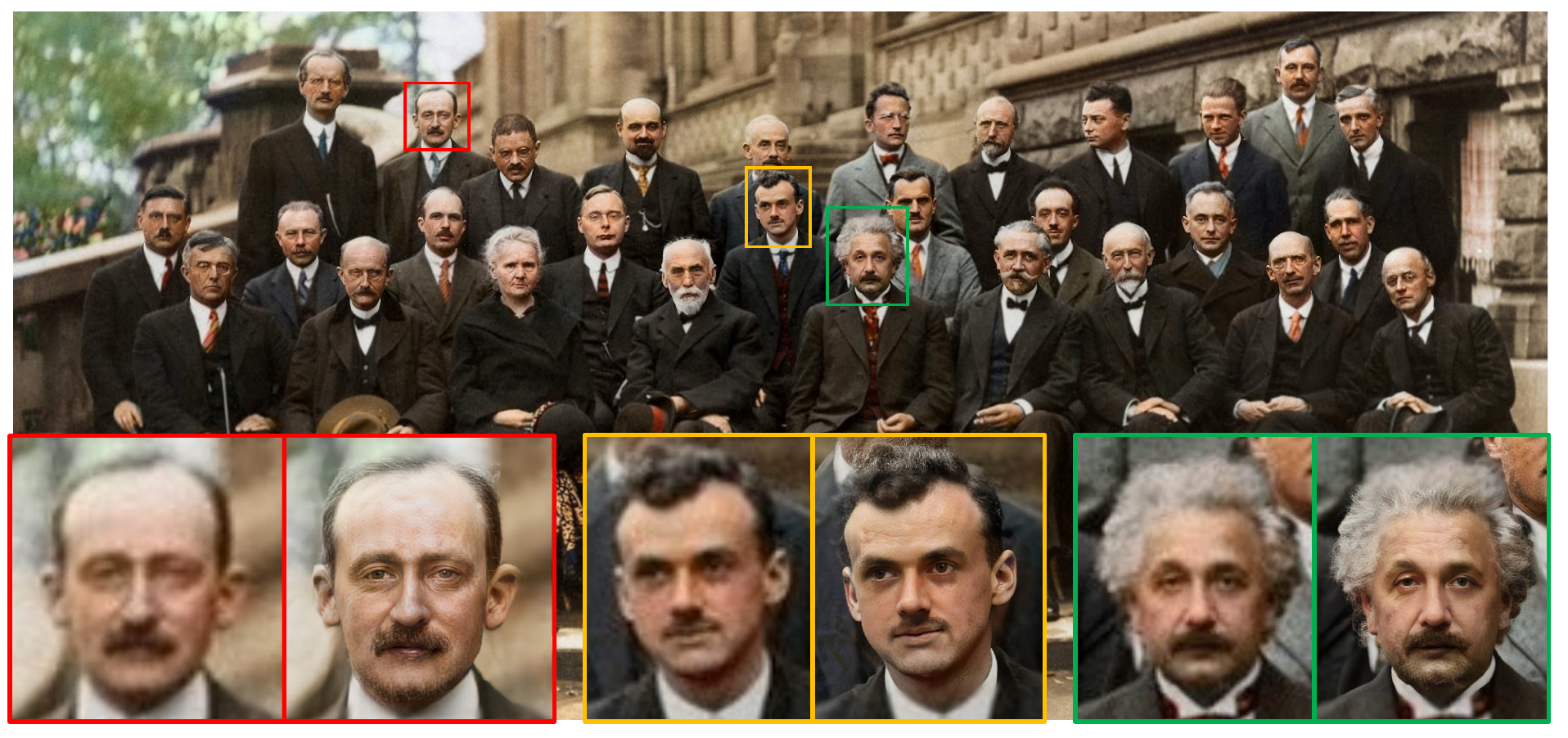}
		\vspace{-0.3cm}
		\caption{\textbf{Results of real-world image restoration.} GLEAN is able to restore the fine details that are not captured in the input image. Image taken at the Solvay conference. Faces in the background image are replaced with restored outputs generated by GLEAN. \textbf{(Zoom in to view other faces in the group photo)}}
		\label{fig:solvay}
		\vspace{-0.5cm}
	\end{center}
\end{figure*}

\vspace{0.15cm}

\noindent{\textbf{Robustness to Poses and Contents.}}
Another appealing property of GLEAN is its robustness to the changes in poses and contents. As shown in Fig.~\ref{fig:sr_pose}, guided by the convolutional features, GLEAN is able to construct realistic non-aligned and non-human face images,  even through the model was trained on aligned human faces. In contrast, the outputs of PULSE are biased to aligned human faces. Its outputs can only approximate the ground truth in low resolution.
Such robustness enables GLEAN to be applied to diverse categories and scenes such as cats, cars, bedrooms, and towers. Examples are shown in Fig.~\ref{fig:sr_others}.

\noindent{\textbf{Quantitative Comparison.}}
To demonstrate the ability of GLEAN in producing outputs with high fidelity,
we extract 100 images from CelebA-HQ~\cite{karras2018progressive} and compute the cosine similarity to the ground-truth on the ArcFace embedding space~\cite{deng2019arcface}. As shown in Table~\ref{tab:arcface}, both GLEAN and LightGLEAN achieve higher similarity than the baseline methods, validating the effectiveness of using generative latent bank in preserving intrinsic structure of the input space.

We additionally provide the quantitative comparison on different categories in Table~\ref{tab:quan}. For each category, we select 100 images and compute their average PSNR and LPIPS~\cite{zhang2018unreasonable}. It is observed that mGANprior and PULSE perform significantly worse as they fail to restore the original objects.
GLEAN outperforms these methods in most categories, suggesting its effectiveness in generating images with high quality and fidelity. In addition, with our effective pruning, LightGLEAN matches the performance of GLEAN and outperforms existing works. Remarkably, LightGLEAN outperforms ESRGAN$^+$ with about 50\% of FLOPs reduction.

\begin{table}[!t]
	\caption{\textbf{Complexity Comparison.} GLEAN and LightGLEAN possess faster speeds with better performance. FLOPs of methods requiring test-time-training are not reported. \textbf{Bolded} texts represent the faster speed.}
	\label{tab:speed}
	\begin{center}
		\tabcolsep=0.15cm
		\vspace{-0.5cm}
		\scalebox{0.85}{
			\begin{tabular}{l|c|c|c|c}
				        & Bicubic                        & PULSE~\cite{menon2020pulse}      & mGANprior~\cite{gu2020image} & DGP~\cite{pang2020exploiting} \\\hline
				Runtime & -                              & 5s                               & 7m                           & 25m                           \\
				Params  & -                              & 30.4M                            & 30.4M                        & 30.4M                         \\
				FLOPs   & -                              & -                                & -                            & -                             \\ \hline\hline
				        & SinGAN~\cite{shaham2019singan} & ESRGAN$^+$~\cite{wang2018esrgan} & \textbf{GLEAN}               & \textbf{LightGLEAN}           \\\hline
				Runtime & 42m                            & 0.13s                            & 0.12s                        & \textbf{0.10s}                \\
				Params  & \textbf{1.03M}                 & 23.97M                           & 175.6M                       & 36.3M                         \\
				FLOPs   & -                              & 225.0G                           & 277.5G                       & \textbf{98.24G}               \\
			\end{tabular}}
	\end{center}
\end{table}
\noindent{\textbf{Complexity Comparison.}} To demonstrate the efficiency of GLEAN and LightGLEAN, we compare their runtime, model size, and FLOPs on the task of $16{\times}$ face super-resolution, using a V100 GPU.
As shown Table~\ref{tab:speed}, GLEAN and LightGLEAN have much faster speeds when compared to GAN-inversion methods and SinGAN. They also outperform ESRGAN$^+$ with comparable speed. Although LightGLEAN possesses a larger model size than existing methods', it has only 44\% of FLOPs of ESRGAN's, and it does not require training during test time.
Despite the large FLOPs improvement of LightGLEAN over the original GLEAN, we observe a minor improvement in runtime reduction due to the non-optimized code. The speed of LightGLEAN could be further improved with more engineering efforts.

\noindent{\textbf{Latent Bank Fintuning.}} Different from DGP~\cite{pang2020exploiting} and GPEN~\cite{yang2021gan}, our experiments show that finetuning the latent bank does not lead to significant performance difference.
The discrepancy results from the differnece in network designs:
(1) DGP represents an image with only latent vectors, which possess limited representational power. Therefore, finetuning is necessary to overfit to the structure of the input image. In contrast, GLEAN uses additional convolutional features to guide the structures.
(2) The encoder features of GPEN are used to replace the \textit{random noise} in the original StyleGAN. Since the noise controls only fine-grained attributes before finetuning, it is expected that finetuning is needed to alter the contribution of the noise. In contrast, GLEAN did not attempt to modify the inputs to StyleGAN.

\subsection{Multi-Class Image Super-Resolution}
In this work, we take one step towards generic image super-resolution by demonstrating the possiblity of leveraging multi-class prior in generative models for multi-class image super-resolution. We employ BigGAN in place of StyleGAN as the latent bank for a multi-class prior.

The quantitative comparisons to existing state of the arts are shown in Table~\ref{tab:quan_multi}, where we obtain the same conclusion in the class-specific image super-resolution. On the one hand, we see that GLEAN and LightGLEAN significantly outperform GAN-inversion methods including PULSE and DGP. On the other hand, by employing the generative prior, our two models also perform favorably against the feedforward model -- ESRGAN. The performance gain is also reflected in the qualitative results depicted in Fig.~\ref{fig:multi_sr}, in which GLEAN produces results with much less artifacts and textures with much better quality. Notably, we observe a larger performance gap between ESRGAN$^+$ and GLEANs in the case of $8{\times}$ super-resolution, underscoring the significance of our generative prior.

\subsection{Image Colorization}
In addition to texture and shape priors that are useful for super-resolution, we hypothesize that various information such as color is captured in the generative model. Therefore, in this work, we extend our notion to a task orthogonal to super-resolution -- colorization, to demonstrate the versatility of GLEAN.
For this task, we also use the multi-class prior in BigGAN. The training scheme remains the same, and the network architecture is modified as follows:
\begin{enumerate}
	\item The input is the luminance channel in the \textit{Lab} color space, and the input channel of the first convolution is modified from $3$ to $1$.
	\item The decoder is replaced with four convolutional layers, and the output channel of the last convolutional layer is $2$. The output is then concatenated to the input luminance image.
\end{enumerate}
More details about the architecture is provided in the supplementary material.

\noindent\textbf{Performance.}
We compare the performance of the proposed method with representative methods including DGP~\cite{pang2020exploiting}, InstColorization~\cite{su2020instance}, and DeOldify~\cite{deoldify}.
Table~\ref{tab:quan_multi} and Fig.~\ref{fig:multi_color} demonstrate the superiority of GLEAN in comparison to existing methods. Following existing works~\cite{su2020instance}, we employ PSNR and LPIPS as the evaluation metrics. From Table~\ref{tab:quan_multi} we see that GLEAN achieves significant gains over existing works, and LightGLEAN is slightly inferior to GLEAN but achieves the second best performance. The qualitative comparison is shown in Fig.~\ref{fig:multi_color}, in which GLEAN outperforms other methods, highlighting the model's capability in capturing color prior. More examples are provided in the supplementary material.

\begin{table}[!t]
	\caption{\textbf{Quantitative (PSNR/LPIPS) comparison on ImageNet, using BigGAN as the latent bank.} GLEAN outperforms other methods in most categories. \textbf{Bolded} texts represent the best performance.}
	\label{tab:quan_multi}
	\begin{center}
		\tabcolsep=0.1cm
		% \vspace{-0.5cm}
		\scalebox{0.9}{
			\begin{tabular}{l|c|c}
				                                 & 4x SR                 & 8x SR                          \\\hline
				PULSE~\cite{menon2020pulse}      & 14.88/0.6923          & 14.80/0.6808                   \\
				DGP~\cite{pang2020exploiting}    & 20.56/0.2818          & 18.69/0.3904                   \\
				ESRGAN$^+$~\cite{wang2018esrgan} & 22.89/0.1442          & 20.04/0.2628                   \\
				\textbf{GLEAN}                   & 23.12/\textbf{0.1239} & \textbf{20.62}/\textbf{0.2388} \\
				\textbf{LightGLEAN}              & \textbf{23.13}/0.1312 & 20.52/0.2541                   \\
			\end{tabular}}
		\tabcolsep=0.1cm
		\scalebox{0.9}{
			\begin{tabular}{l|c}
				                                 & Colorization                   \\\hline
				DGP~\cite{pang2020exploiting}    & 21.34/0.1950                   \\
				InstColor.~\cite{su2020instance} & 22.88/0.1807                   \\
				DeOldify~\cite{deoldify}         & 23.12/0.1609                   \\
				\textbf{GLEAN}                   & \textbf{23.48}/\textbf{0.1469} \\
				\textbf{LightGLEAN}              & 23.24/0.1534                   \\
			\end{tabular}}
	\end{center}
\end{table}

\begin{table}[!t]
	\caption{\textbf{Quantitative (NIQE/FID/Identity similarity) comparison on real-world face image restoration.} GLEAN outperforms existing state of the arts, producing outputs of high quality and fidelity. \textbf{Bolded} texts represent the best performance.}
	\label{tab:blind}
	\vspace{-0.5cm}
	\begin{center}
		\tabcolsep=0.1cm
		\scalebox{1}{
			\begin{tabular}{l|c|c}
				                                    & LFW\_a~\cite{huang2008labeled}       & Celeb-A~\cite{liu2015deep}           \\\hline
				PULSE~\cite{menon2020pulse}         & 5.018/82.10/0.2509                   & 3.709/83.02/0.3084                   \\
				DFDNet~\cite{li2020blind}           & 8.878/82.35/0.8584                   & 10.152/73.01/0.8780                  \\
				PSFR-GAN~\cite{chen2021progressive} & 5.999/69.87/0.8466                   & 5.989/65.83/0.8945                   \\
				\textbf{GLEAN}                      & 3.943/\textbf{67.71}/\textbf{0.8949} & 3.921/\textbf{61.65}/\textbf{0.9192} \\
				\textbf{LightGLEAN}                 & \textbf{3.578}/68.67/0.8596          & \textbf{3.635}/62.35/0.8774          \\
			\end{tabular}}
		\vspace{-0.5cm}
	\end{center}
\end{table}
\noindent\textbf{Discussion.}
Both existing works and GLEANs attempt to use various forms of priors to improve the perceptual quality of colorization results.
In particular, DGP and DeOldify both adopt adversarial loss to better approximate the natural image manifold.
InstColorization employs an off-the-shelf detection network to provide object prior. With the knowledge of the object category, the search space of the color could be reduced.
Our method explores another use of prior. Through training for image generation, the network is required to understand the color distribution of the respective category in order to synthesize realistic objects. GLEAN exploits this information to synthesize realistic color for various objects.
\subsection{Real-World Face Image Restoration}
In this section, we demonstrate the capability of GLEAN for real-world face image restoration. In the original GLEAN, bicubic downsampling is added during training. As a result, GLEAN cannot be applied to real-world images with unknown degradations. To further improve the generalizability of GLEAN, we apply random degradations during training to mimic the complex degradations in reality. In this task, the ill-poseness increases significantly and the use of priors becomes more critical.
We first describe the training and test settings, followed by the comparison with existing state of the arts. The training scheme remains unchanged.\\

\noindent\textbf{Training Data.}
We train GLEAN on synthetic data that approximate real low-quality images.
We follow the degradation process adopted in ~\cite{li2020blind} for synthesizing the training data:
\begin{equation}\label{equ:degradation}
	y = \left[(x\circledast k_{\sigma})_{\downarrow_{r}} + n_{\delta}\right]_{\mathtt{JPEG}_{q}}.
\end{equation}
In other words, the high quality image $x$ is first convolved with a Gaussian blur kernel $k_{\sigma}$ followed by a downsampling operation with a scale factor $r$. After that, additive white Gaussian noise $n_{\delta}$ is added to the image, and finally the noisy image is compressed by JPEG with quality factor $q$.
For each training pair, we randomly sample values of $\sigma$, $r$, $\delta$, and $q$ from the intervals $[0.2,10]$, $[1,8]$, $[0,25]$, and $[5,50]$, respectively.\\

\noindent\textbf{Test Data and Metrics.} We use two existing face image datasets for testing. LFW\_a is a subset of the LFW dataset~\cite{huang2008labeled} whose elements have a surname starting with the letter ``A''. In addition, we select the first 500 images (images with IDs from 1 to 500) from Celeb-A~\cite{liu2015deep} for test. Since no ground-truth images are available, we use NIQE~\cite{mittal2013making}, FID~\cite{heusel2017gans}, and ArcFace similarity~\cite{deng2019arcface} as the evaluation metrics.\\

\noindent\textbf{Qualitative Comparison.}
The qualitative comparison is shown in Fig.~\ref{fig:blind}. By optimizing only the low-dimensional vectors, PULSE~\cite{menon2020pulse} successfully reconstructs natural images, but the outputs are dissimilar to the ground truths. When compared with DFDNet~\cite{li2020blind} and PFSRGAN~\cite{chen2021progressive}, which are the current state-of-the-art blind face image restoration methods, GLEAN is able to produce more details and hence images with better quality.\\

\noindent\textbf{Quantitative Comparison.}
The quantitative comparison is shown in Table~\ref{tab:blind}.
First, we compare the image naturalness using NIQE and FID. GLEAN and LightGLEAN outperform existing methods in these metrics, demonstrating the superiority of these two methods in the generation of high-quality images.
Second, we compare the image fidelity using the ArcFace similarity. We observe that the outputs of PULSE has a significantly worse similarity, indicating that PULSE severely alters the identity of the inputs. Our models achieve either similar to or better similarity in both datasets, confirming its effectiveness in preserving identity.

To further demonstrate its effectiveness on real-world face restoration, we apply GLEAN on a group photo taken at the Solvay conference. As illustrated in Fig.~\ref{fig:solvay}, GLEAN successfully restores the fine details that are not captured in the original images.

\begin{figure}[!t]
	\begin{center}
		\hspace{-0.5cm}\includegraphics[width=0.45\textwidth]{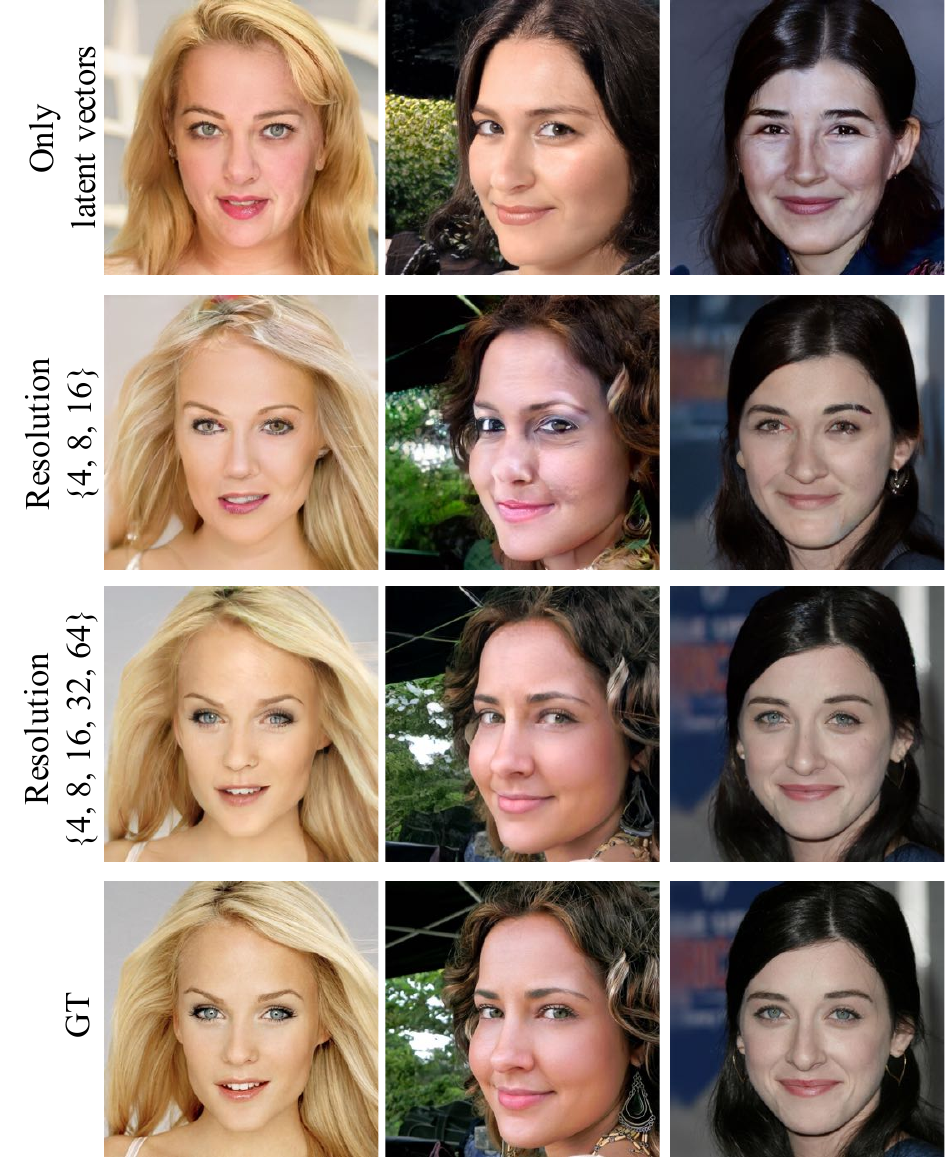}
		\caption{\textbf{Effects of the multi-resolution encoder features.} Without the convolutional features, the outputs can only resemble the global attributes (\eg~hair color, pose). When adding the encoder features progressively, the network can capture more local structures, better approximating the GT. }
		\label{fig:ablation_enc}
		\vspace{-0.4cm}
	\end{center}
\end{figure}

% !TEX root = ../submission.tex
\begin{figure}[!t]
    \begin{center}
        \includegraphics[width=0.45\textwidth]{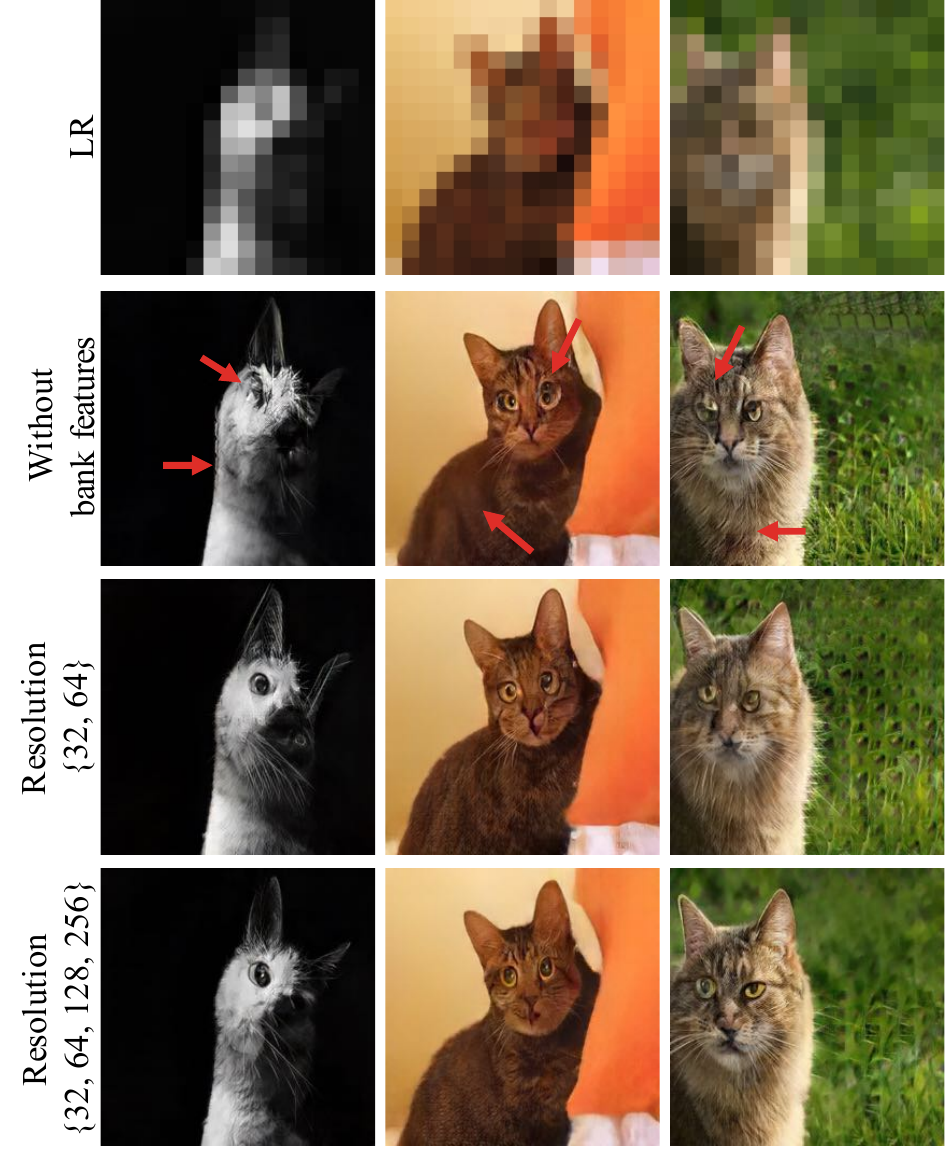}
        \caption{\textbf{Effects of the latent bank features.} The rich texture priors captured in the generator lift the burden of the encoder in texture generation. Improvements on both texture and structures are observed when finer features are inserted into the decoder. \textbf{(Zoom in for best view)}}
        \label{fig:ablation_gen}
    \end{center}
\end{figure}
\begin{figure}[!t]
    \begin{center}
        \includegraphics[width=0.45\textwidth]{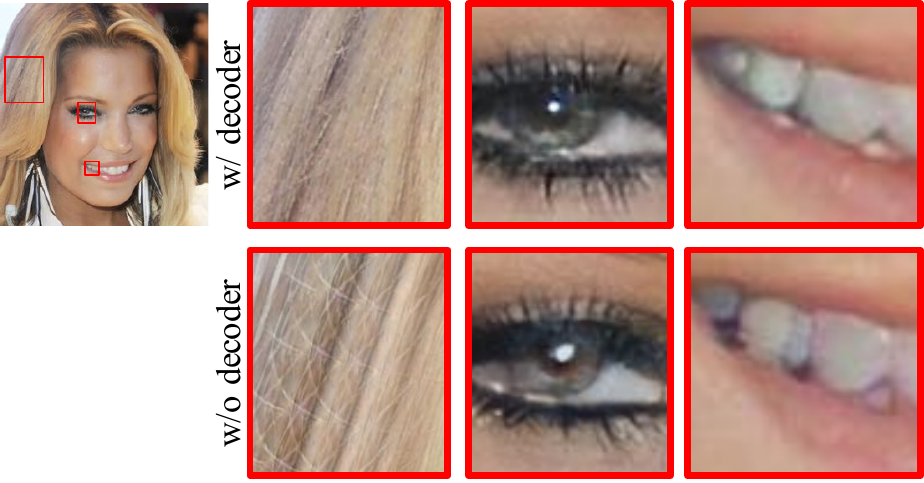}
        \caption{\textbf{Contributions of the decoder.} The decoder reinforces the spatial information captured in the encoder features and aggregate them in a coarse-to-fine manner, resulting in an enhanced quality.}
        \vspace{-0.5cm}
        \label{fig:ablation_dec}
    \end{center}
\end{figure}

\section{Ablation Studies}
\label{sec:ablation_enc}
\noindent\textbf{Importance of Multi-resolution Encoder Features.}
We demonstrate how the convolutional features generated from the encoder assist in the restoration of fine details and local structures.
We start with only the latent vectors and observe the transition when features are gradually introduced to the latent bank as conditions.
To remove the effects brought on by the decoder, we test with a variant of GLEAN where the generator directly produces the output images. The comparison is depicted in Fig.~\ref{fig:ablation_enc}.

When all convolutional features are discarded, GLEAN resembles the typical GAN inversion methods that learn only the latent vectors. Similar to those methods, the network is able to synthesize realistic images given the latent vectors. However, guided only by low-dimensional vectors, which spatial information is not well-preserved, the network restores only the global attributes such as hair color and poses, but fails to preserve finer details.
When providing coarse (from $4{\times}4$ to $16{\times}16$) convolutional features to the latent bank, more details are recovered, and the outputs are better approximations of the ground truths. Further improvements in both quality and fidelity are observed when finer features are passed to the latent bank.
The above observations corroborate our hypothesis that the convolutional features are pivotal in guiding the restoration of fine details and local structures, which cannot be reconstructed with only the latent vectors.

\vspace{0.15cm}
\noindent\textbf{Effects of Latent Bank Features.}
To understand the contributions of the latent bank, we investigate the effects brought on by the latent bank features. We start by discarding all the latent bank features and progressively pass the features to the decoder. The comparison is shown in Fig.~\ref{fig:ablation_gen}.
Lacking appropriate prior information, the network is responsible for both generating realistic details and maintaining fidelity to the ground-truths. Such a demanding objective eventually leads to outputs that contain flaws in both structure restoration and texture generation. 
With the latent bank, the burden of texture and details generation is reduced as the generator already captures rich image priors. Therefore, improvements in both structures and textures are observed when passing finer features to the decoder.

\vspace{0.15cm}
\noindent\textbf{Importance of Decoder.}
As shown in Fig.~\ref{fig:ablation_dec}, without the decoder, despite being perceptually convincing overall, the output image contains unpleasant artifacts when zoomed in.
The decoder allows the network to aggregate the information in a coarse-to-fine manner, leading to more natural details. In addition, the multi-scale skip-connections between the encoder and decoder reinforce the spatial information captured in the encoder features so that the latent bank could focus more on detail generation, further enhancing the output quality.

\begin{figure}[!t]
    \begin{center}
        \subfloat[Comparison with DFDNet~\cite{li2020blind}]{\includegraphics[width=0.4\textwidth]{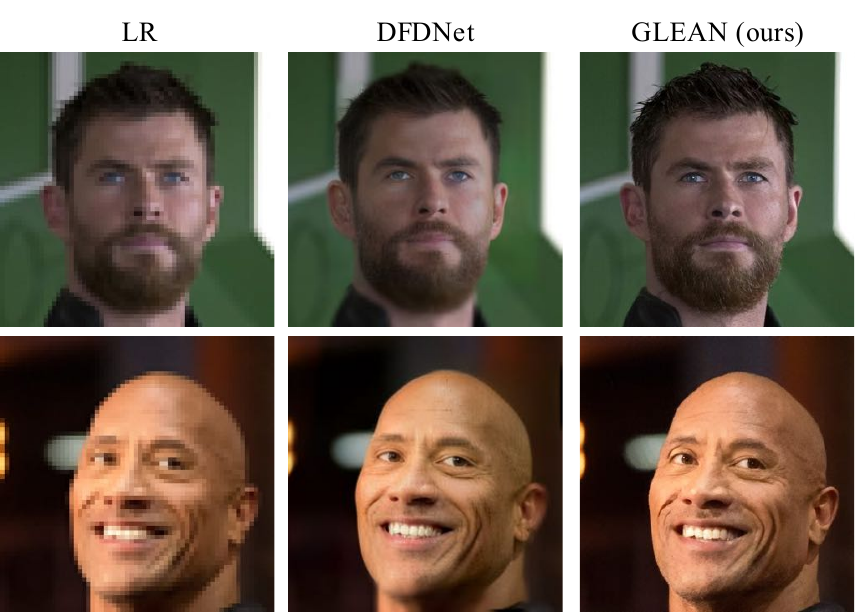}}\\
        \subfloat[Comparison with SRNTT~\cite{zhang2019image}]{\includegraphics[width=0.4\textwidth]{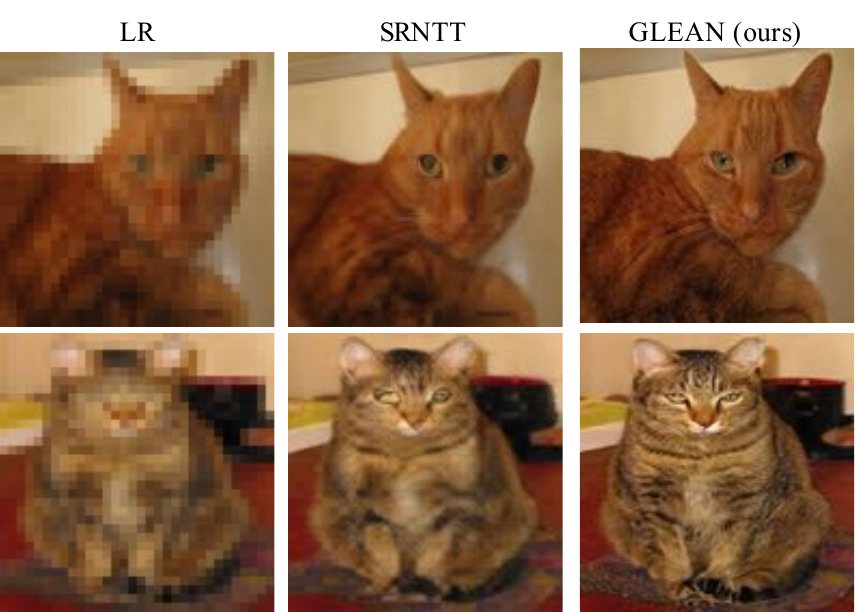}}
        \caption{\textbf{Comparison with imagery dictionary.} \textbf{(a)} DFDNet fails to restore components absent in the dictionary (\eg~skin, hair), leading to incoherent outputs. \textbf{(b)} SRNTT is unable to faithfully produce fur textures.}
        \label{fig:ablation_dict}
        \vspace{-0.4cm}
    \end{center}
\end{figure}
\vspace{0.15cm}
\noindent\textbf{Comparisons with Reference-Based Methods.}
We assess the efficacy of the new notion of GAN-based dictionary by comparing GLEAN with two representative methods that adopt an imagery dictionary for SR -- DFDNet~\cite{li2020blind} and SRNTT~\cite{zhang2019image}. Examples are shown in Fig.~\ref{fig:ablation_dict}.

For DFDNet, we evaluate the performance on LR images with unknown degradations\footnote{We further downsample the LR images to $64{\times}64$ to match the input size of GLEAN.}. Through pre-constructing a dictionary of facial components (\eg~eyes, lips), DFDNet shows remarkable performance on face restoration. However, it cannot faithfully produce results on parts absent in the dictionary, such as skin and hair. Therefore, significant incoherence is observed in the outputs.
Despite GLEAN being trained on the bicubic kernel, it is still capable of producing visually appealing outputs. More importantly, GLEAN is not confined to improving the visual quality of specific components. Instead, the entire image is super-resolved, leading to coherent and attractive results. We foresee GLEAN to achieve even better performance by employing multiple degradations during training

For SRNTT, we follow the same settings and downsample the ground-truth images using the bicubic kernel. With such low-resolution images ($32{\times}32$), global matching becomes prohibitive, and hence SRNTT fails to transfer the textures from HR reference images. As a result, SRNTT tends to provide blurry textures. By capturing the distribution instead of the specific imagery clues, GLEAN does not rely on explicit textural transferal procedure. This enables its applicability to large-factor SR, where image matching is extremely difficult.
More importantly, with no external images employed, GLEAN does not require any global matching to search for suitable textures/details. This allows GLEAN to be applied to images with higher resolutions, where global matching is computationally prohibitive.

% !TEX root = ../submission.tex
\begin{figure}[!t]
    \begin{center}
        \includegraphics[width=0.49\textwidth]{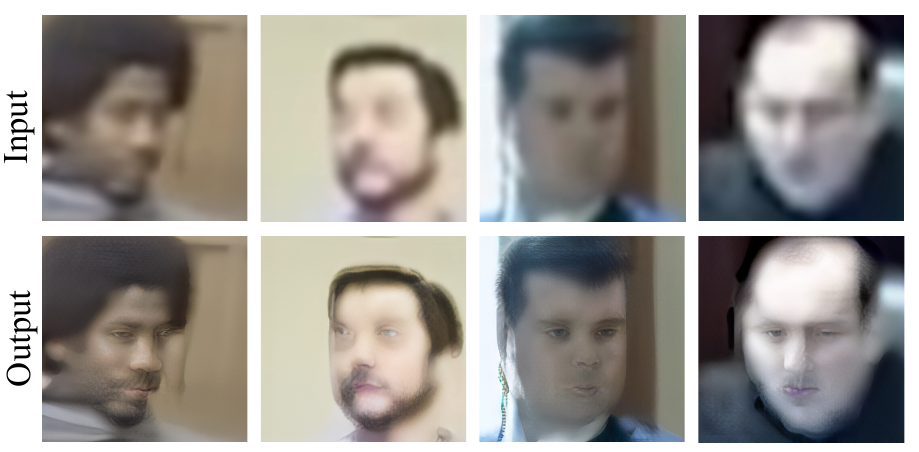}
        \vspace{-0.5cm}
        \caption{\textbf{Failure cases on extreme degradations.} In extreme cases that contain severe degrdations, GLEAN cannot restore extreme cases of severe degradations. This problem can potentially be solved by adopting heavier degradations.}
        \vspace{-0.5cm}
        \label{fig:failure}
    \end{center}
\end{figure}

\section{Discussion and Conclusion}
We have presented a new way to exploit pre-trained GANs for various image restoration tasks including super-resolution, colorization, and hybrid restoration. We have shown that a pre-trained GAN can be used as a generative latent bank in an encoder-bank-decoder architecture.
We have also presented a lightweight version of GLEAN, named LightGLEAN, and demonstrated the potential of these two methods on generic images by employing multi-class prior.
Both GLEAN and LightGLEAN outperform existing state of the arts in terms of fidelity and quality.

Despite obtaining satisfactory results in various class domains, there are a few limitations in GLEANs. First, in spite of the efforts made in this work towards generic restoration, the existing version of GLEANs are still confined to \textit{finite classes} and \textit{fixed resolution} encompassed by the GAN prior, due to the lack of powerful generic image synthesizer. We believe that with a stronger generative model that can synthesize realistic images on more diverse scenes, the performance of GLEANs and relevant ideas could be markedly improved.
Second, in the task of real-world face restoration, although they work well on cases of mild to moderate degradations, they fail to produce pleasant outputs in cases of heavy degradations, as shown in Fig.~\ref{fig:failure}. To adapt to such complex and heavy degradations, we believe that incorporating heavier degradations during training could partially remedy the situation. In addition, more sophisticated data augmentation schemes, such as second-order degradations~\cite{wang2021real} and degradation-shuffling~\cite{zhang2021designing}, could be taken into consideration.
In the future, with more sophisticated generative models, we believe that the notion of generative latent bank can be used for more restoration tasks as well as more diverse scenes. This idea can potentially be extended to various forms of priors such as language priors.

\vspace{0.1cm}
\noindent\textbf{Acknowledgement.}
This study is supported under the RIE2020 Industry Alignment Fund Industry Collaboration Projects (IAF-ICP) Funding Initiative, as well as cash and in-kind contribution from the industry partner(s). It is also partly supported by the NTU NAP grant.

\vspace{-0.1cm}
% Bibliography
\bibliographystyle{IEEEtran}
\bibliography{short,papers}
% \includegraphics command within an optiodai2019secondordernal argument. (You could create

\begin{IEEEbiography}[{\includegraphics[width=1in,height=1.25in,clip,keepaspectratio]{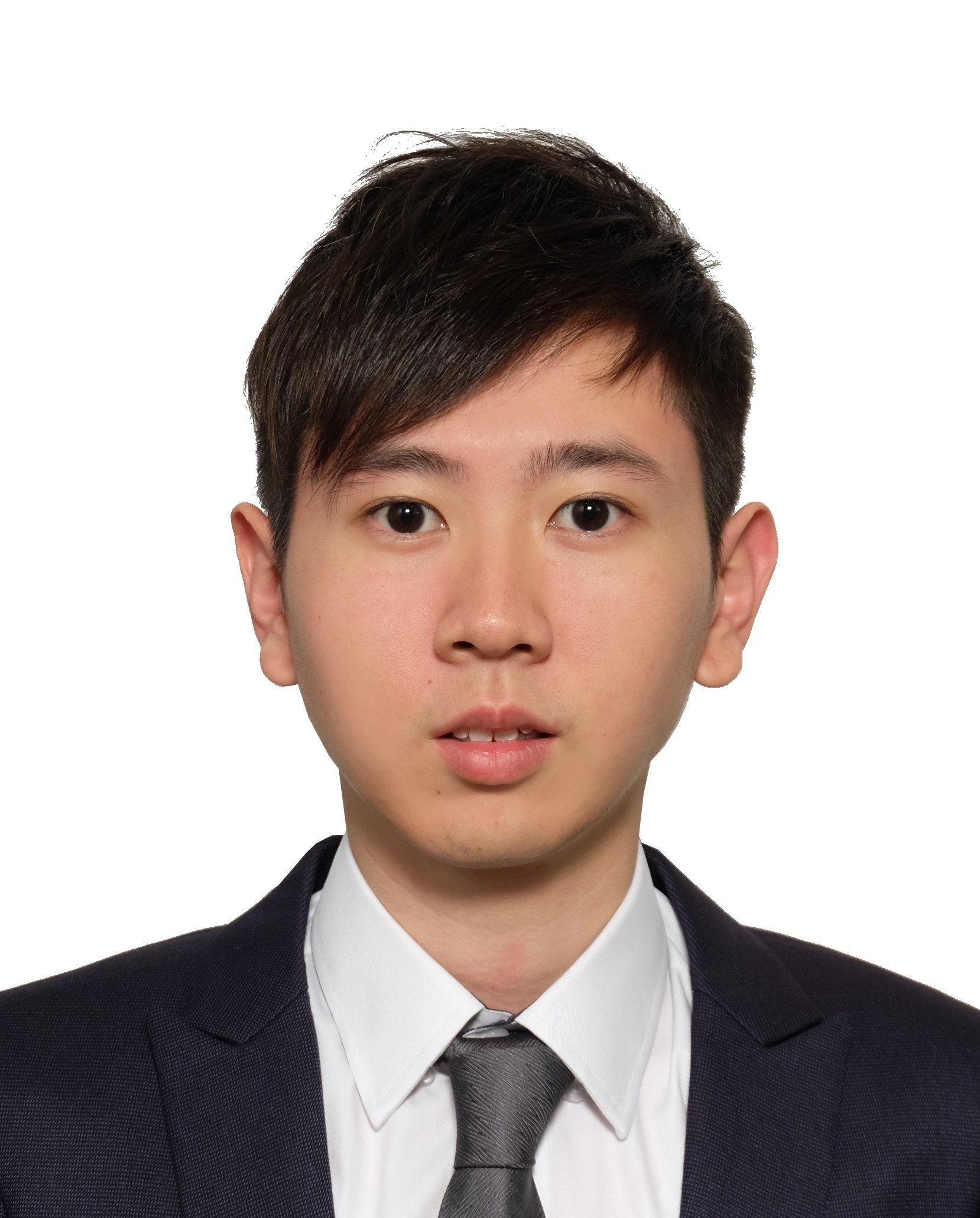}}]{Kelvin C.K. Chan} is currently a fourth-year PhD student at S-Lab, Nanyang Technological University. He received his MPhil degree in Mathematics as well as his BSc and BEng degrees from The Chinese University of Hong Kong. He was awarded the Google PhD Fellowship in 2021. He won the first place in multiple international challenges including NTIRE2019 and NTIRE2021, and was selected as an outstanding reviewer in ICCV 2021. His research interests include low-level vision, especially image and video restoration.
\end{IEEEbiography}

\begin{IEEEbiography}[{\includegraphics[width=1in,height=1.25in,clip,keepaspectratio]{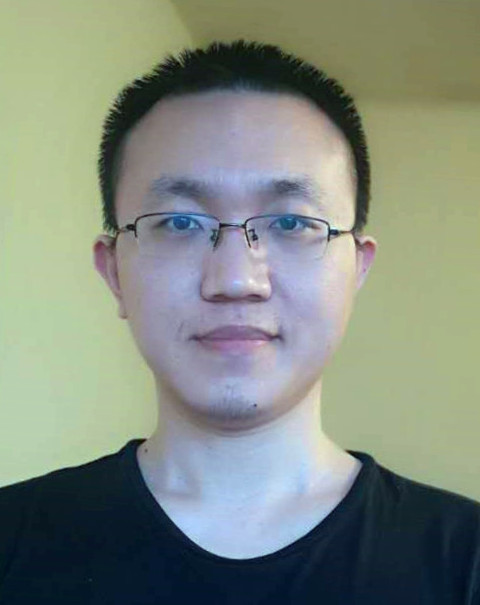}}]{Xiangyu Xu} is a research fellow at S-Lab, Nanyang Technological University. He was a postdoc fellow in the Robotics Institute, Carnegie Mellon University, Pittsburgh, PA, USA from 2019 to 2020, and a research scientist at SenseTime, Beijing, China from 2018 to 2019. He got the Ph.D. degree in the Department of Electronic Engineering, Tsinghua University in 2018. Before that, he received the B.Eng degree in the Department of Electronic Engineering, Tsinghua University in 2013, and the B.Ec degree in the National School of Development, Peking University in 2015.	He was a visiting Ph.D. student at University of California, Merced and Harvard University.	His research interest includes image processing, computer vision, and machine learning.
\end{IEEEbiography}

\begin{IEEEbiography}[{\includegraphics[width=1in,height=1.25in,clip,keepaspectratio]{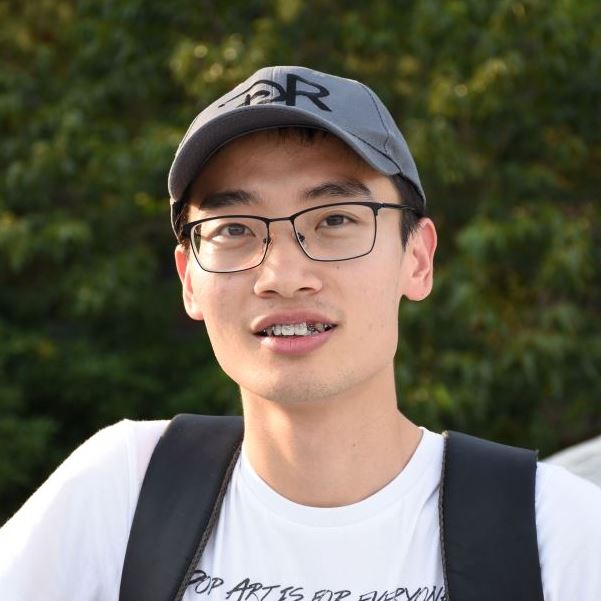}}]{Xintao Wang} is currently a researcher in Applied Research Center (ARC), Tencent PCG. He received his Ph.D. degree in the Department of Information Engineering, The Chinese University of Hong Kong, in 2020. He was selected as an outstanding reviewer in CVPR 2019 and an outstanding reviewer (honorable mention) in BMVC 2019. He won the first place in several international super-resolution challenges including NTIRE2019, NTIRE2018, and PIRM2018. His research interests focus on low-level vision problems, including super-resolution, image and video restoration.
\end{IEEEbiography}

%\newpage

\begin{IEEEbiography}[{\includegraphics[width=1in,height=1.4in,clip,keepaspectratio]{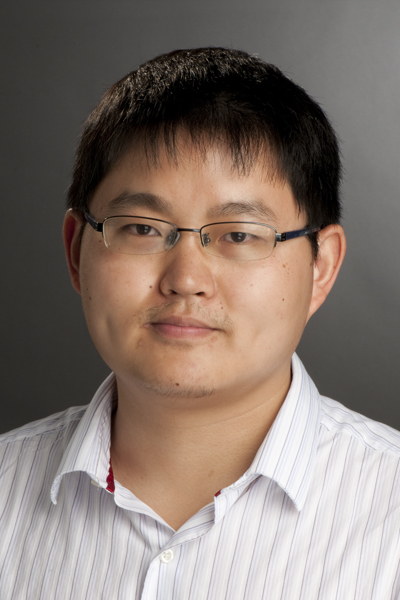}}]{Jinwei Gu} (Senior Member, IEEE) is the R\&D Executive Director of SenseTime USA. His current research focuses on low-level computer vision, computational photography, smart visual sensing and perception, and robotics. He obtained his PhD degree in 2010 from Columbia University, and the B.S and M.S. from Tsinghua University, in 2002 and 2005 respectively. Before joining SenseTime, he was a senior research scientist in NVIDIA Research from 2015 to 2018.  Prior to that, he was an assistant professor in Rochester Institute of Technology from 2010 to 2013, and a senior researcher in the media lab of Futurewei Technologies from 2013 to 2015. He serves as an associate editor for IEEE Transactions on Computational Imaging and IEEE Transactions on Pattern Analysis and Machine Intelligence. He is an IEEE Senior Member since 2018.
\end{IEEEbiography}

\begin{IEEEbiography}[{\includegraphics[width=1in,height=1.25in,clip,keepaspectratio]{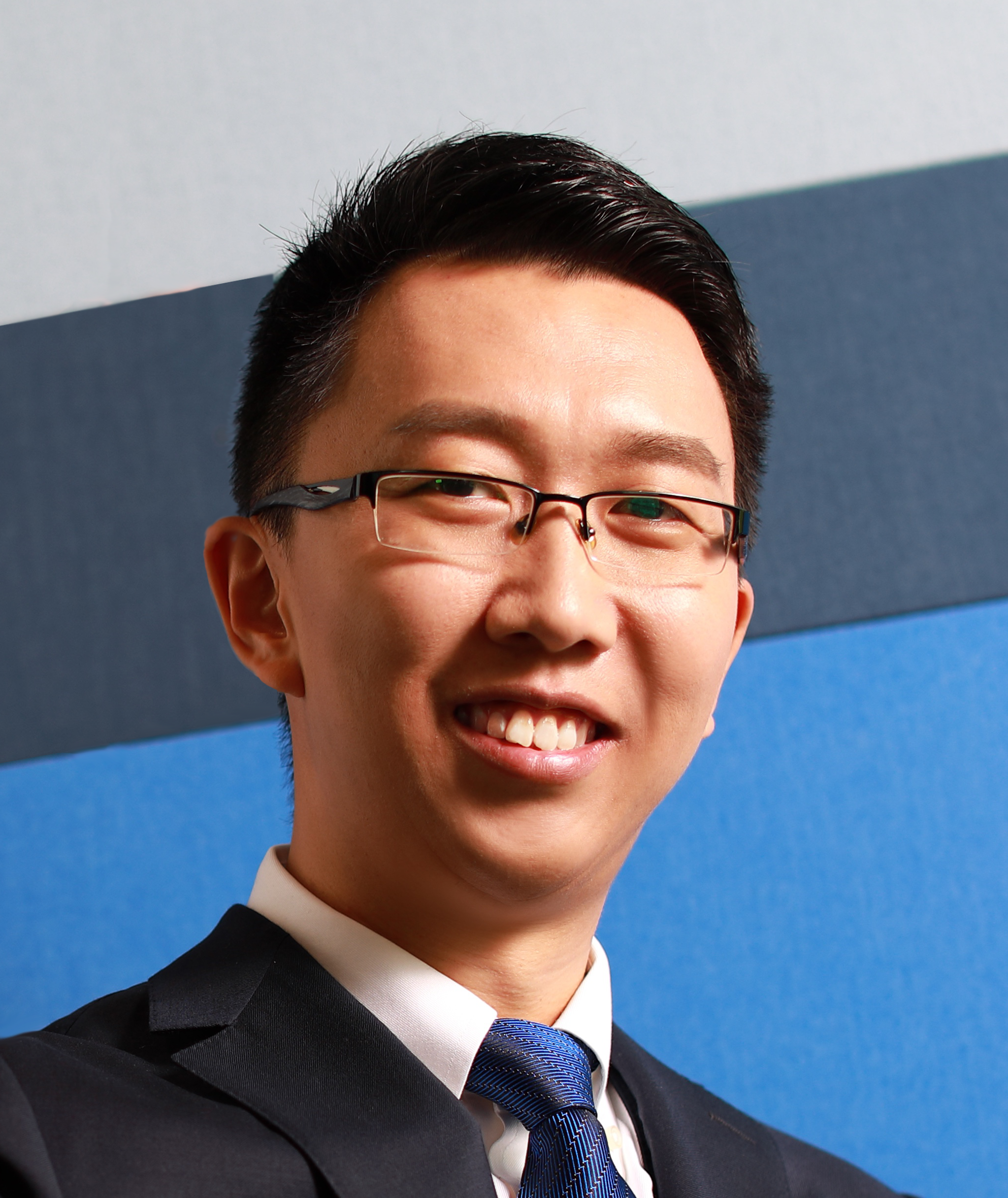}}]{Chen Change Loy} (Senior Member, IEEE) is an Associate Professor with the School of Computer Science and Engineering, Nanyang Technological University, Singapore. He is also an Adjunct Associate Professor at The Chinese University of Hong Kong. He received his Ph.D. (2010) in Computer Science from the Queen Mary University of London. Prior to joining NTU, he served as a Research Assistant Professor at the MMLab of The Chinese University of Hong Kong, from 2013 to 2018. He was a postdoctoral researcher at Queen Mary University of London and Vision Semantics Limited, from 2010 to 2013.
  He serves as an Associate Editor of the IEEE Transactions on Pattern Analysis and Machine Intelligence and   International Journal of Computer Vision. He also serves/served as an Area Chair of major conferences such as ICCV 2021, CVPR 2021, CVPR 2019, and ECCV 2018. He is a senior member of IEEE.
  His research interests include image/video restoration and enhancement, generative tasks, and representation learning.
\end{IEEEbiography}


%\vfill


% !TEX root = ../submission.tex
\clearpage
\onecolumn
\appendix
\appendices
We will first discuss the change of the architecture we have made for the task of colorization in Sec.~\ref{sec:arch}. We will then provide additional qualitative results of GLEAN and LightGLEAN in Sec.~\ref{sec:results}.

\section{Architecture for Colorization}
\label{sec:arch}
Unlike super-resolution, in the task of colorization, the resolution of the input image is the same as that of the output image. Therefore, we made minimal modifications to the architecture so that GLEANs can be adapted to the case when input and output have the same resolution. The network architecture is modified as follows:
\begin{enumerate}
	\item The input is the luminance channel in the \textit{Lab} color space, and the input channel of the first convolution is modified from $3$ to $1$.
	\item The decoder is replaced with four convolution layers, and the output channel of the last convolutional layer is $2$. The output is then concatenated to the input luminance image.
\end{enumerate}
We find that this simple modification suffices to achieve good colorization results. Overall, the architectures for different tasks are highly similar, demonstrating the versatility of GLEAN and LightGLEAN.

\section{Qualitative Results}
\label{sec:results}
\begin{figure*}[!b]
	\begin{center}
		\includegraphics[width=\textwidth]{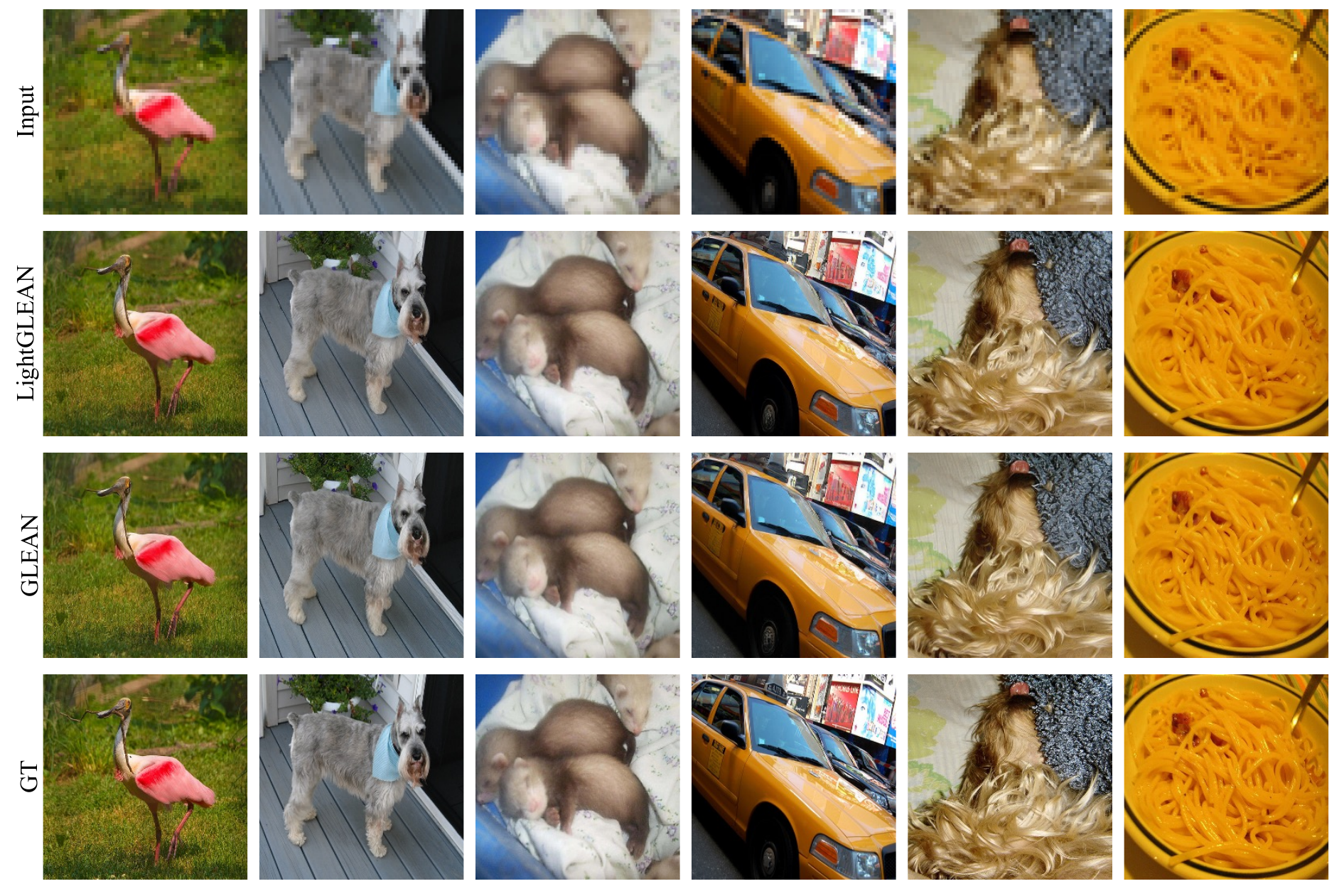}
		\caption{More results of GLEAN and LightGLEAN on $4{\times}$ multi-class image super-resolution. With the powerful priors, GLEAN and LightGLEAN are able to produce realistic details. \textbf{(Zoom in for best view)}}
		\label{fig:supp_4x}
	\end{center}
\end{figure*}

\begin{figure*}[!t]
	\begin{center}
		\includegraphics[width=\textwidth]{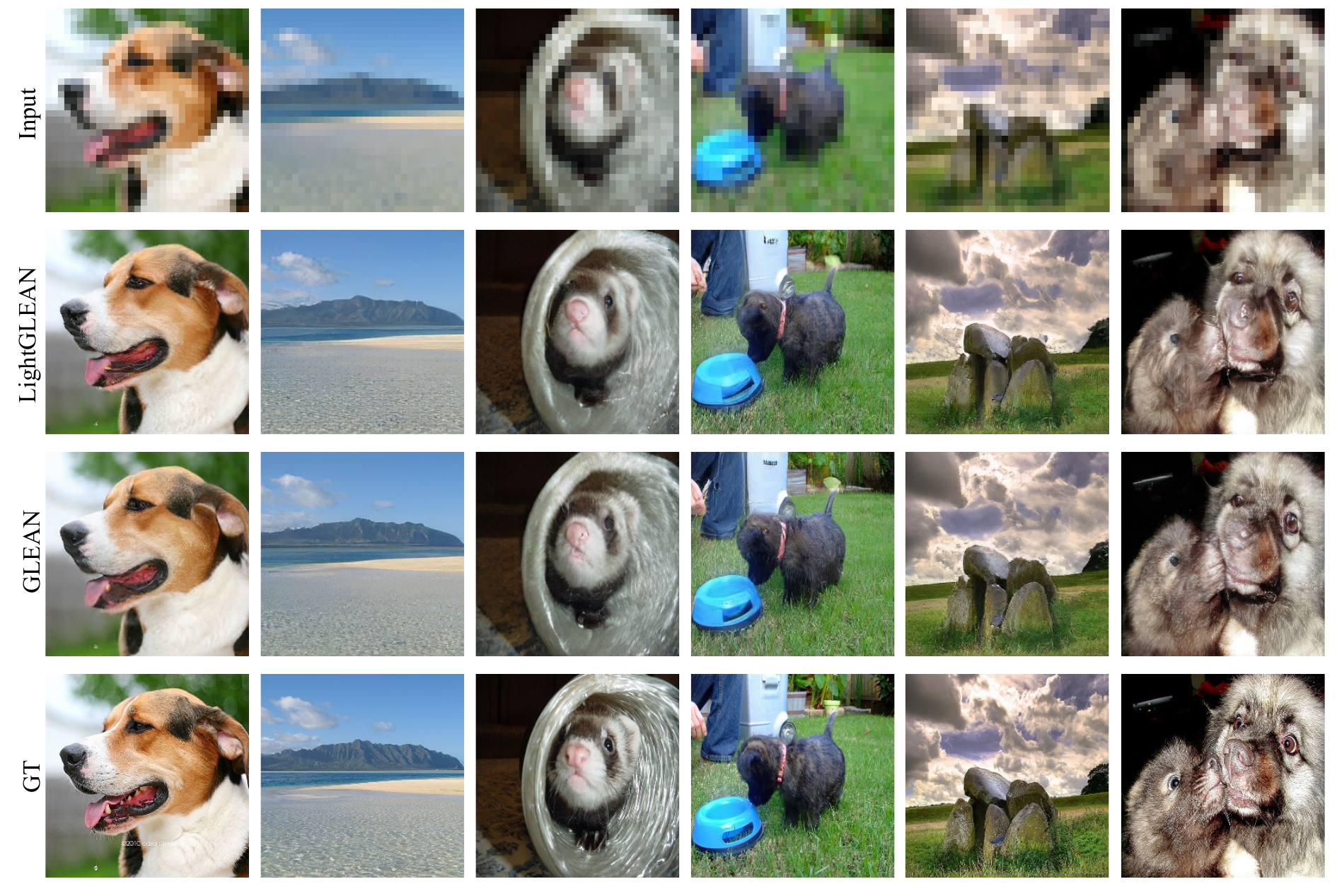}
		\caption{More results of GLEAN and LightGLEAN on $8{\times}$ multi-class image super-resolution. Despite the input image is only $32{\times}32$, our proposed models are able to synthesize realistic textures. \textbf{(Zoom in for best view)}}
		\label{fig:supp_8x}
	\end{center}
\end{figure*}

\begin{figure*}[!t]
	\begin{center}
		\includegraphics[width=\textwidth]{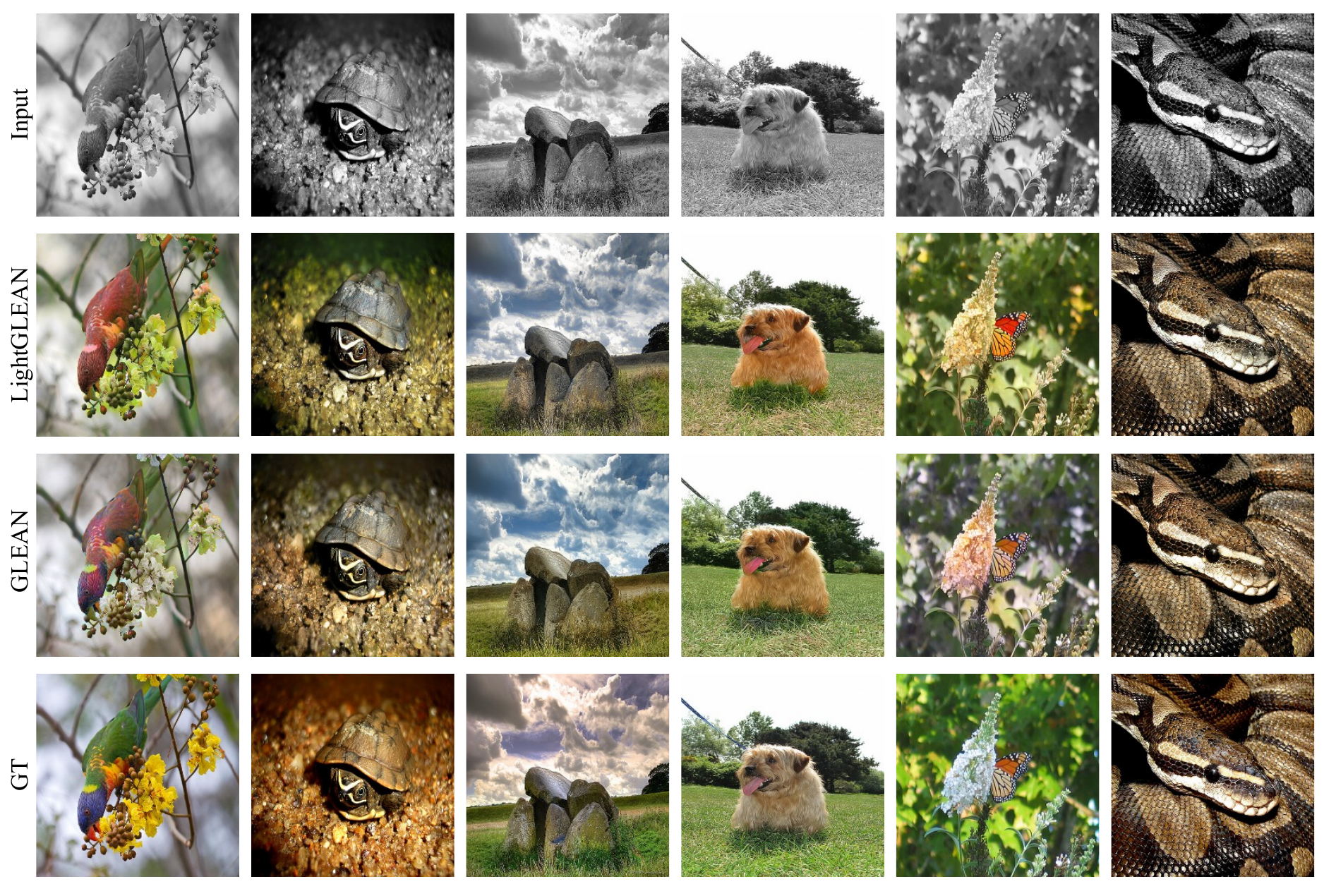}
		\caption{More results of GLEAN and LightGLEAN on multi-class image colorization. Although the color is sometimes dissimilar to the ground-truths, our models are able to produce natural color.}
		\label{fig:supp_color_2}
	\end{center}
\end{figure*}

\begin{figure*}[!t]
	\begin{center}
		\includegraphics[width=\textwidth]{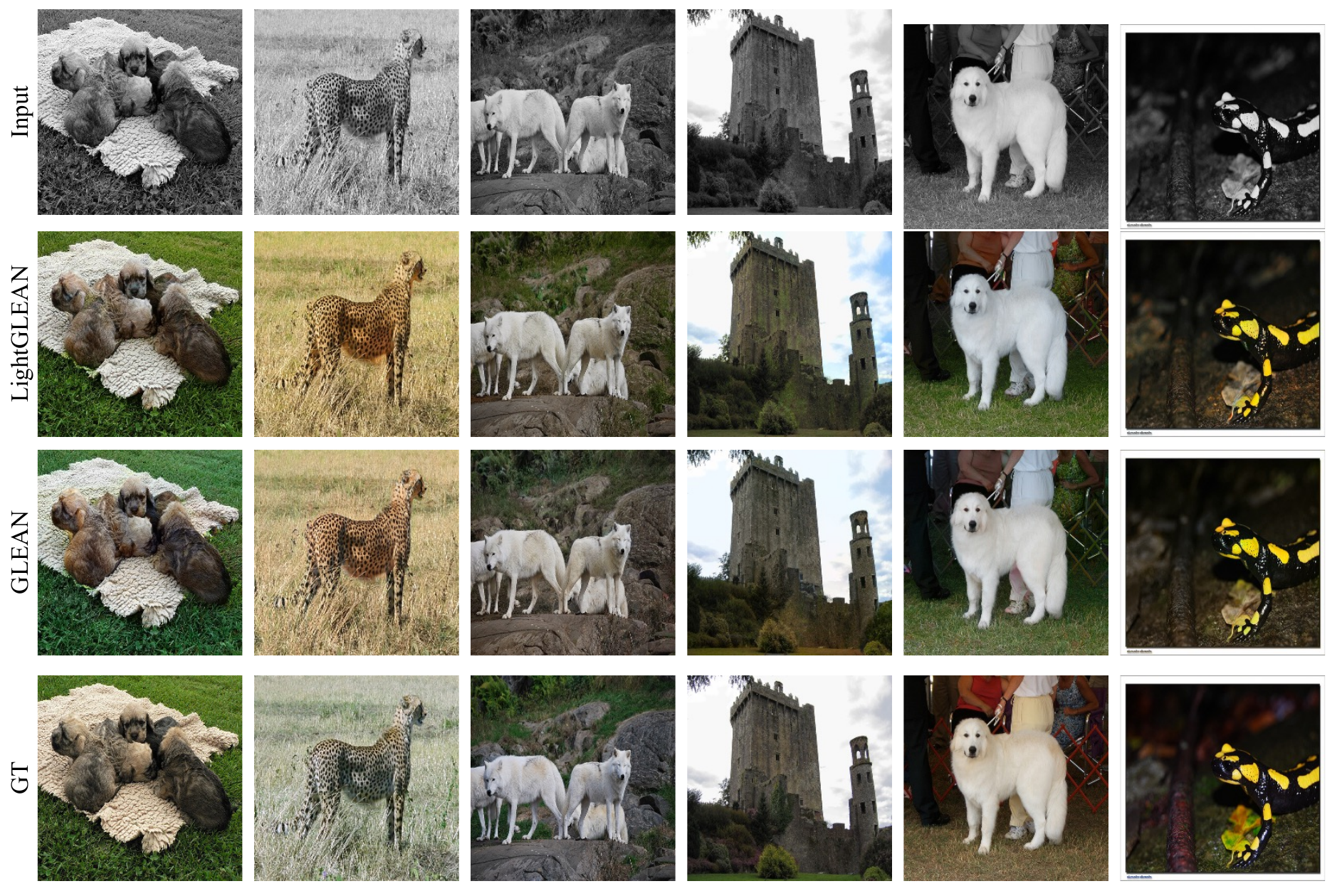}
		\caption{More results of GLEAN and LightGLEAN on multi-class image colorization. Although the color is sometimes dissimilar to the ground-truths, our models are able to produce natural color.}
		\label{fig:supp_color}
	\end{center}
\end{figure*}

\begin{figure*}[!t]
	\begin{center}
		\includegraphics[width=\textwidth]{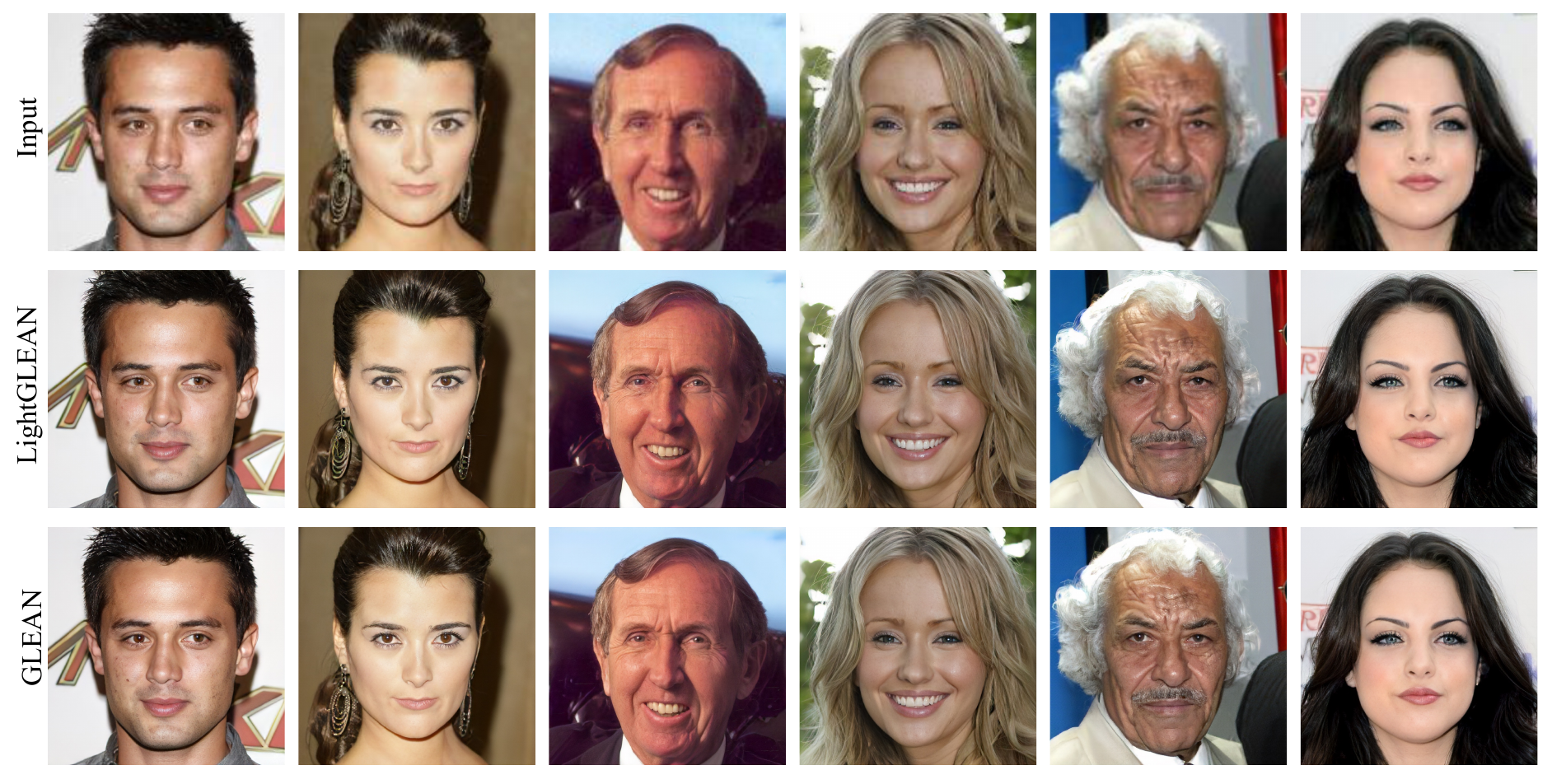}
		\caption{More results of GLEAN and LightGLEAN on real-world face image restoration. Our models are able to restore real-world face images, which contain diverse and unknown degradations. Note that the ground-truth is unavailable for real-world images. \textbf{(Zoom in for best view)}}
		\label{fig:supp_real}
	\end{center}
\end{figure*}

In this section, we demonstrate additional qualitative results on 1) multi-class image super-resolution, 2) image colorization, and 3) real-world face image restoration. From Fig.~\ref{fig:supp_4x} to Fig.~\ref{fig:supp_real} we observed that with the generative priors encapsulated in our latent bank, our methods are able to produce faithful results despite the highly ill-posed nature of the tasks.

\end{document}